%% file: main.tex
\documentclass[11pt]{article}

\usepackage[final]{acl}

\usepackage{times}
\usepackage{latexsym}
\usepackage[T1]{fontenc}
\usepackage[utf8]{inputenc}
\usepackage{microtype}
\usepackage{graphicx}

\usepackage{amsmath}
\usepackage{amssymb}
\usepackage{booktabs}
\usepackage{multirow}
\usepackage{makecell}
\usepackage{array}
\usepackage{xcolor}
\usepackage{xspace}
\usepackage[inline]{enumitem}
\usepackage{colortbl}

\hypersetup{
  pdftitle={OSPD: On-Policy Self-Distillation for Persona-Consistent Dialogue},
  pdfauthor={Rui Xu, Yikai Zhang, Aili Chen, Zicheng Zhao, Xu Yinghui, Libo Wu}
}

\newcommand{\eg}{\textit{e.g.}\xspace}
\newcommand{\wo}{\textit{w/o}\xspace}
\newcommand{\etc}{\textit{etc.}\xspace}

\newcommand{\method}{\textsc{OSPD}\xspace}
\newcommand{\cfull}{$c_\text{full}$\xspace}
\newcommand{\cbrief}{$c_\text{brief}$\xspace}

\title{OSPD: On-Policy Self-Distillation\\for Persona-Consistent Dialogue}

\author{
\textbf{Rui Xu}$^{1,2}$ \quad \textbf{Yikai Zhang}$^{1}$ \quad \textbf{Aili Chen}$^{1}$ \\
\textbf{Zicheng Zhao}$^{1}$ \quad \textbf{Xu Yinghui}$^{1}$ \quad \textbf{Libo Wu}$^{1,2}$ \\
$^{1}$Fudan University \\
$^{2}$Shanghai Innovation Institute \\
\texttt{24110240097@m.fudan.edu.cn}
}

\begin{document}
\maketitle

\begin{abstract}
\input{Sections/0_Abstract}
\end{abstract}

\begin{figure}[!t]
\centering
\includegraphics[width=\columnwidth]{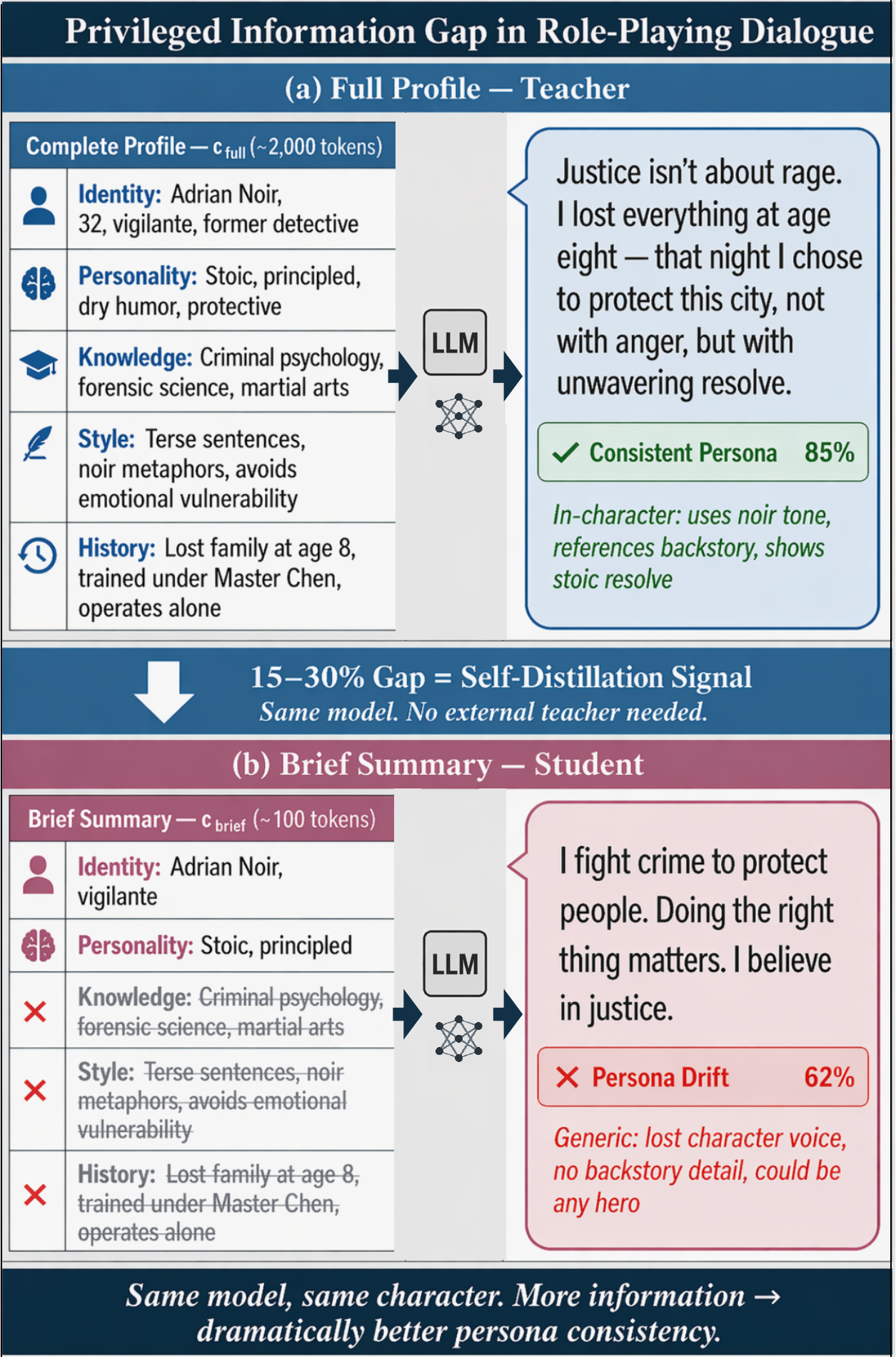}
\caption{Privileged information gap in role-playing dialogue.
The same model conditioned on a complete character profile (\cfull) produces substantially more persona-consistent responses than when conditioned on a brief summary (\cbrief).
This gap provides a natural self-distillation signal without any external teacher.}
\label{fig:motivation}
\end{figure}

\input{Sections/1_Introduction}
\input{Sections/2_Related_Work}
\input{Sections/3_Method}
\input{Sections/4_Experiments}
\input{Sections/5_Analysis}
\input{Sections/6_Conclusion}

\section*{Limitations}
\input{Sections/Limitations}

\section*{Ethics Statement}
\input{Sections/Ethics}

\section*{Acknowledgments}
We thank the anonymous reviewers and meta-reviewer for their constructive feedback.
During camera-ready preparation, we used OpenAI Codex to assist with formatting, code review, and language editing.
The authors reviewed all suggested changes and remain fully responsible for the paper and released artifacts.

\bibliography{custom}

\appendix
\input{Sections/Appendix}

\end{document}

%% file: Sections/0_Abstract.tex
Maintaining persona consistency across multi-turn dialogues remains a core challenge for role-playing language models.
Off-policy distillation from external teachers incurs distribution mismatch that compounds across dialogue turns, while reinforcement learning struggles with reward ambiguity inherent in subjective persona fidelity.
We propose \method, an on-policy self-distillation framework where the same model serves as both teacher and student under asymmetric information: the teacher receives a complete character profile while the student sees only a brief summary, and the student generates trajectories from its own policy.
We find that teacher confidence in role-playing dialogue exhibits a bimodal structure---sharply peaked at character-critical tokens yet diffuse at generic utterances---and introduce role-aware divergence switching to match this structure.
A progressive trait masking curriculum further forces staged internalization of character knowledge along semantic dimensions.
Experiments on CharacterBench, CharacterEval, and SocialBench show that \method substantially improves persona consistency over supervised fine-tuning and multi-turn RL baselines, without requiring any external teacher or reward model.

%% file: Sections/1_Introduction.tex
\section{Introduction}
\label{sec:intro}

Maintaining a coherent persona across extended dialogue is a central challenge for role-playing conversational agents, with applications in interactive entertainment, educational simulation, and social companionship~\citep{pmlr-v267-wang25dk,yu-etal-2025-beyond,NEURIPS2025_4c914438}.
A well-functioning agent must faithfully reflect a character's personality, knowledge, and behavioral patterns throughout a conversation~\citep{wang-etal-2024-incharacter}---yet current systems suffer from \textit{persona drift}: the model gradually abandons its assigned character and reverts to generic behavior as the dialogue progresses~\citep{NEURIPS2025_4c914438}.
From an imitation-learning perspective, this is a manifestation of \textit{exposure bias}~\citep{pmlr-v15-ross11a}---errors compound across turns because the model generates from its own policy at inference but trains on external data, and a single out-of-character reply corrupts the entire subsequent context.

Existing methods attack this problem through contrastive learning~\citep{ji-etal-2025-enhancing}, reinforcement learning~\citep{NEURIPS2025_4c914438,wang2026verirole,ye-etal-2025-cpo}, and reasoning augmentation~\citep{NEURIPS2025_aacca7b6,qin-etal-2025-r}, but each carries fundamental limitations.
Off-policy distillation incurs distribution mismatch whose cumulative error grows quadratically with dialogue length~\citep{pmlr-v15-ross11a}.
RL methods require carefully engineered rewards for the inherently subjective notion of persona fidelity.
Contrastive approaches struggle with fine-grained trait distinctions.
No prior work addresses persona consistency through on-policy self-distillation, despite its theoretical advantages for reducing exposure bias.

We propose \method, an on-policy self-distillation framework that exploits two properties of role-playing dialogue (Figure~\ref{fig:framework}).
\textbf{First}, the same model shows a large persona-consistency gap under different information budgets (Figure~\ref{fig:motivation}): responses conditioned on a complete character profile (\cfull, $\sim$2\,000 tokens) are substantially more consistent than those conditioned on a brief summary (\cbrief, $\sim$100 tokens).
This gap yields a natural self-distillation signal---the model's own \textit{privileged-information mode} serves as the teacher, eliminating the need for any external model.
\textbf{Second}, teacher confidence follows a \textit{bimodal} distribution: sharply peaked at character-critical tokens (\eg catchphrases, value judgments) yet diffuse at generic utterances.
This structure motivates \textit{role-aware divergence switching}---reverse KL at low-entropy positions for precise character imitation, forward KL elsewhere to preserve diversity.
\method further introduces a \textit{progressive trait masking} curriculum that incrementally withholds character knowledge from the student across training, forcing the model to internalize persona information into its parameters rather than relying on the input prompt.

Experiments on CharacterBench~\citep{zhou2024characterbenchbenchmarkingcharactercustomization}, CharacterEval~\citep{tu-etal-2024-charactereval}, and SocialBench~\citep{chen-etal-2024-socialbench} demonstrate that \method substantially outperforms supervised fine-tuning and is competitive with or superior to multi-turn RL baselines, while requiring neither an external teacher nor a reward model.
Multi-turn analysis shows markedly slower persona drift as dialogue length increases to 60 turns.
Ablations confirm the contribution of each component, and an analysis of the learned divergence gate validates the bimodal structure across diverse characters and topics.
Code and configurations are available at \url{https://github.com/airaer1998/OSPD}.

Our contributions are threefold.
\begin{enumerate*}[label=\textit{\arabic*)}]
\item We introduce on-policy self-distillation with privileged information to role-playing dialogue, eliminating the need for external teachers or reward models.
\item We identify a bimodal confidence structure specific to role-playing dialogue and design a role-aware divergence switching mechanism that exploits it.
\item We propose progressive trait masking, a curriculum that forces staged internalization of character knowledge along semantic dimensions, and validate it with systematic experiments including ablations, multi-turn robustness analysis, and divergence-gate interpretability studies.
\end{enumerate*}

%% file: Sections/2_Related_Work.tex
\section{Related Work}
\label{sec:related}

\paragraph{Knowledge Distillation for Language Models.}
On-policy distillation lets the student generate its own training trajectories, eliminating off-policy distribution mismatch.
GKD~\citep{ICLR2024_5be69a58}, MiniLLM~\citep{ICLR2024_8ac015d4}, and DistiLLM~\citep{pmlr-v235-ko24c} can be viewed as instances of a unified $f$-divergence objective~\citep{song2026surveyonpolicydistillationlarge}.
Recent work introduces per-token adaptive divergences: ToDi~\citep{jung-etal-2025-todi} mixes forward and reverse KL by teacher--student logit ratios, and Entropy-Aware OPD~\citep{jin2026entropyawareonpolicydistillationlanguage} gates selection by teacher entropy.
Orthogonally, self-distillation removes external teacher dependencies: SPIN~\citep{pmlr-v235-chen24j} distinguishes model-generated from human text via self-play, while OPSD~\citep{zhao2026selfdistilledreasoneronpolicyselfdistillation}, OPCD~\citep{ye2026onpolicycontextdistillationlanguage}, and GATES~\citep{stein2026gatesselfdistillationprivilegedcontext} use privileged context as the supervision signal.
\citet{penaloza2026privilegedinformationdistillationlanguage} formalize conditions for privileged information distillation.
Existing applications focus on reasoning and instruction following, where the privileged signal is a single correct answer; our work targets role-playing dialogue, where the privileged information is a multi-dimensional character profile whose informativeness varies across positions---motivating both our adaptive divergence and trait masking curriculum.

\paragraph{Persona-Consistent Role-Playing.}
Methods for persona consistency include contrastive learning~\citep{ji-etal-2025-enhancing}, multi-turn RL~\citep{NEURIPS2025_4c914438,wang2026verirole,ye-etal-2025-cpo}, and reasoning augmentation~\citep{NEURIPS2025_aacca7b6,qin-etal-2025-r}.
On the data side, CoSER~\citep{pmlr-v267-wang25dk} provides authentic literary dialogues with structured character knowledge, and Beyond Dialogue~\citep{yu-etal-2025-beyond} introduces profile--dialogue alignment.
Beyond dialogue fidelity, recent work studies persona-driven decisions~\citep{xu-etal-2025-character}, role-playing agents for key opinion leaders~\citep{xu2024mindechoroleplayinglanguageagents}, and inner-thought reasoning~\citep{xu-etal-2025-guess}.
None of these methods exploit the information asymmetry between detailed and condensed character profiles, nor do they leverage on-policy self-distillation to internalize character knowledge.

%% file: Sections/3_Method.tex
\section{Method}
\label{sec:method}

\subsection{Problem Formulation}
\label{sec:formulation}

A character is represented as a pair $\mathcal{C} = (c_\text{full}, c_\text{brief})$, where $c_\text{full}$ ($\sim$2\,000 tokens) is a \textit{privileged} profile containing detailed background, psychological portrait, relationship graph, representative dialogues, and linguistic style, and $c_\text{brief}$ ($\sim$100 tokens) is a condensed summary retaining only name and core personality traits.
At inference time, only $c_\text{brief}$ is available---this reflects the practical setting where users provide a short character description rather than an exhaustive dossier.

Let $h_t = (u_1, a_1, \ldots, u_t)$ denote the dialogue history up to turn~$t$.
We define two modes of the \textit{same} model:
\begin{align}
\pi_T(a \mid c_\text{full}, h_t) &= \text{LM}_{\bar{\theta}}(a \mid c_\text{full}, h_t) \label{eq:teacher} \\
\pi_S(a \mid c_\text{brief}, h_t) &= \text{LM}_{\theta}(a \mid c_\text{brief}, h_t) \label{eq:student}
\end{align}
where $\theta$ are the student parameters being optimized and $\bar{\theta}$ is an exponential moving average (EMA) of $\theta$.
The information asymmetry resides entirely in the conditioning context; the EMA copy provides a slowly evolving target that stabilizes training (\S\ref{sec:ema}).

\input{Figures/framework_placeholder}

Figure~\ref{fig:framework} illustrates the overall pipeline.
The student generates dialogue trajectories from its own policy, the teacher evaluates each trajectory under privileged information, and the resulting token-level distributional signal drives parameter updates through a role-aware divergence objective.

\subsection{On-Policy Self-Distillation}
\label{sec:on_policy}

To eliminate the distribution mismatch inherent in off-policy distillation, the student generates its own training trajectories.
Each training step proceeds as follows.

\paragraph{Step 1: On-Policy Generation.}
The student samples a response $\hat{a}_t \sim \pi_S(\cdot \mid c_\text{brief}, h_t)$.
The training distribution is the student's own behavioral distribution, directly removing exposure bias.

\paragraph{Step 2: Teacher Evaluation.}
Using the same dialogue context $h_t$ but conditioned on $c_\text{full}$, the EMA teacher computes a distribution over each token position of $\hat{a}_t$:
\begin{equation}
\pi_T(\cdot \mid c_\text{full}, h_t, \hat{a}_{t,<k}) \quad \text{for } k = 1, \ldots, |\hat{a}_t|
\label{eq:teacher_eval}
\end{equation}
These distributions serve as position-wise supervision targets.

\paragraph{Step 3: Divergence Computation.}
The role-aware divergence loss $\mathcal{L}_\text{OSPD}$ is computed over the trajectory (\S\ref{sec:divergence}).

\paragraph{Step 4: Gradient Update.}
Gradients flow only through the student's log-probabilities; teacher logits are treated as fixed targets.

A key distinction from pure self-play methods such as SPIN~\citep{pmlr-v235-chen24j} is that the \method teacher continuously injects external information through $c_\text{full}$.
The model does not self-reinforce its own biases; instead, it aligns toward a distribution conditioned on richer information, preventing the policy saturation that plagues closed-loop self-distillation.

\paragraph{EMA Teacher Stabilization.}
\label{sec:ema}
Sharing parameters between teacher and student creates a moving-target problem: naive parameter sharing causes the target distribution to shift with every gradient step, risking training oscillation.
We maintain an EMA copy of the student parameters as the teacher:
\begin{equation}
\bar{\theta} \leftarrow \alpha\, \bar{\theta} + (1 - \alpha)\, \theta
\label{eq:ema}
\end{equation}
with momentum $\alpha \in [0.99, 0.999]$, following the mean-teacher paradigm~\citep{NIPS2017_68053af2}.
The EMA teacher evolves slowly, providing a smoothed target that lags behind the student by a controlled margin.

\subsection{Role-Aware Divergence Switching}
\label{sec:divergence}

The bimodal confidence structure of role-playing dialogue necessitates position-dependent learning strategies.
When the teacher processes character-critical content---domain knowledge, catchphrases, value judgments---the privileged profile provides decisive guidance and the teacher distribution concentrates sharply (low entropy).
At generic utterances---greetings, transitions, open-ended remarks---even full profile access leaves considerable uncertainty (high entropy).

We compute the teacher entropy at each token position~$k$:
\begin{equation}
\mathcal{H}_k = \mathcal{H}\bigl(\pi_T(\cdot \mid c_\text{full}, h, \hat{a}_{<k})\bigr)
\label{eq:entropy}
\end{equation}
and define a gating coefficient:
\begin{equation}
\alpha_k = \sigma\!\left(\frac{\mathcal{H}_k - \mu_{\mathcal{H}}}{\tau}\right)
\label{eq:gate}
\end{equation}
where $\mu_{\mathcal{H}}$ is the running mean of teacher entropy within the current batch and $\tau$ is a temperature hyperparameter.

The distillation loss interpolates between reverse and forward KL:
\begin{equation}
\begin{split}
\mathcal{L}_\text{OSPD} = \sum_{k}\Bigl[&(1 - \alpha_k)\, D_\text{KL}(\pi_S \| \pi_T) \\
  &+ \alpha_k\, D_\text{KL}(\pi_T \| \pi_S)\Bigr]
\end{split}
\label{eq:loss_ospd}
\end{equation}
Low $\alpha_k$ (high teacher confidence) activates \textbf{reverse KL}, which is mode-seeking and drives precise imitation of peaked character behaviors.
High $\alpha_k$ (low teacher confidence) activates \textbf{forward KL}, which is mode-covering and preserves expressive diversity.

This mechanism differs from generic adaptive divergence methods~\citep{jung-etal-2025-todi,jin2026entropyawareonpolicydistillationlanguage} in a fundamental way: our switching logic is grounded in the semantic structure of role-playing dialogue.
Teacher entropy directly reflects the informativeness of the privileged profile at the current position---low entropy signals that $c_\text{full}$ provides decisive guidance, high entropy signals that the profile is non-informative here.
The divergence choice thus functions as an implicit detector of \textit{where privileged information matters}.

\subsection{Progressive Trait Masking Curriculum}
\label{sec:masking}

Naive information reduction---simply shortening the profile---ignores the hierarchical structure of character knowledge.
We decompose the character profile into five semantic dimensions:
\begin{enumerate*}[label=\textit{\roman*)}]
\item \textbf{Identity}: name, age, occupation, relationships;
\item \textbf{Personality}: core values, emotional disposition;
\item \textbf{Knowledge}: domain expertise, character-specific facts;
\item \textbf{Style}: speech patterns, vocabulary, register;
\item \textbf{History}: formative events, growth arc.
\end{enumerate*}

Training proceeds in two stages:

\paragraph{Stage 1 (Warm-up, 1 epoch).}
The student receives a compressed version of all five dimensions ($\sim$500 tokens), while the teacher accesses the full profile.
The moderate information gap eases early optimization.

\paragraph{Stage 2 (Masking, 2 epochs).}
The student sees only \textit{Identity} and \textit{Personality} ($\sim$100 tokens); \textit{Knowledge}, \textit{Style}, and \textit{History} are masked.
The teacher retains the full profile.
This forces the model to internalize deeper character knowledge into its parameters rather than relying on the input context.

The masking order reflects the information hierarchy at inference time: identity and personality constitute ``public'' information that users always specify, whereas knowledge, style, and history are ``deep'' information that is difficult to convey in a short prompt yet critical for persona consistency.
Stage~1 uses an elevated temperature~$\tau$ to broaden the gate $\alpha_k$ toward forward KL, providing broad-coverage learning; Stage~2 restores the standard temperature, letting the divergence switch be fully entropy-driven.

\subsection{Trajectory Filtering}
\label{sec:pcv}

Not all on-policy trajectories are useful for distillation.
When the student generates a severely out-of-character response, the teacher's distribution conditioned on this aberrant prefix may itself become unreliable.
We introduce a lightweight \textit{Persona Consistency Verifier} (PCV) that filters trajectories at two granularities: turn-level (trait adherence of each reply) and dialogue-level (cross-turn consistency).
Only accepted trajectories contribute to $\mathcal{L}_\text{OSPD}$.

The PCV is an off-the-shelf quality filter, \textit{not} a training objective---\method does not optimize for PCV scores, which fundamentally distinguishes it from RL approaches that optimize against a reward model.
Our primary implementation uses GPT-5.1 as the LLM judge; Appendix~\ref{app:pcv_variants} compares it with an embedding-based MLP classifier, which yields a similar CharacterBench average (3.48 vs.\ 3.51).

\subsection{Training Objective}
\label{sec:objective}

The overall loss combines the self-distillation objective with a supervised fine-tuning (SFT) regularizer:
\begin{equation}
\mathcal{L} = \mathcal{L}_\text{OSPD} + \lambda \cdot \mathcal{L}_\text{SFT}
\label{eq:total_loss}
\end{equation}
where $\mathcal{L}_\text{SFT}$ is the standard next-token prediction loss on curated character dialogues and $\lambda$ is tuned from $\{0.1, 0.5, 1.0\}$.
The SFT term maintains generation quality and prevents the mode collapse that can arise from pure distributional matching.

%% file: Figures/framework_placeholder.tex
\begin{figure*}[t]
\centering
\includegraphics[width=0.95\textwidth]{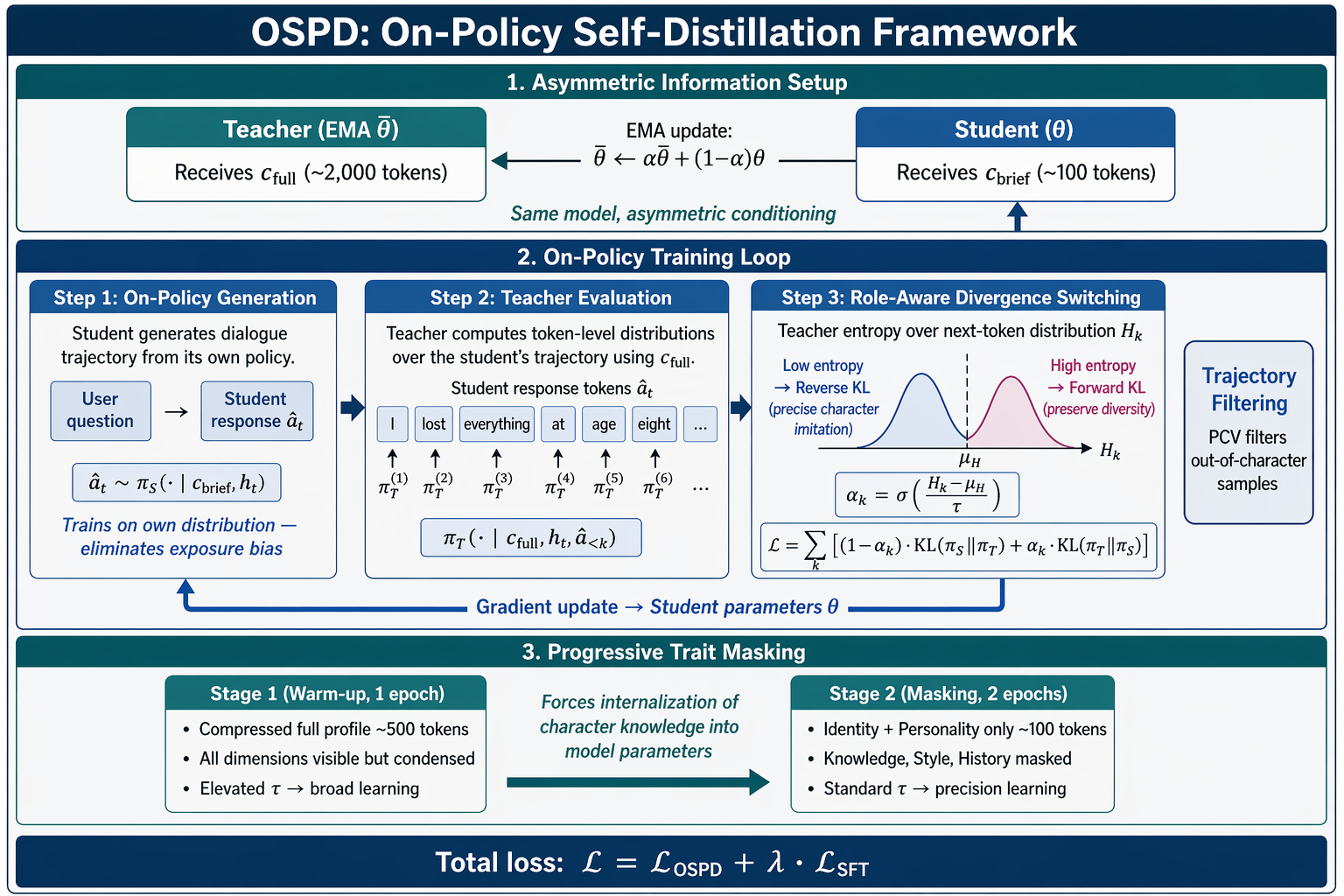}
\caption{Overview of \method.
The same model serves as both teacher and student under asymmetric information conditions.
The student generates on-policy trajectories from $c_\text{brief}$; the EMA teacher provides token-level supervision from $c_\text{full}$.
Role-aware divergence switching adapts the distillation objective to the bimodal confidence structure of role-playing dialogue.
Progressive trait masking transitions from a compressed full profile (Stage~1) to identity and personality only (Stage~2).}
\label{fig:framework}
\end{figure*}

%% file: Sections/4_Experiments.tex
\section{Experiments}
\label{sec:experiments}

\subsection{Setup}
\label{sec:setup}

\paragraph{Models and Training.}
We use Qwen2.5-7B-Instruct as the base model for all experiments and extend to Qwen2.5-14B-Instruct for scaling analysis (Appendix~\ref{app:scaling}).
All methods are fine-tuned with LoRA (rank\,=\,64, $\alpha$\,=\,16, dropout\,=\,0.1) using AdamW with a learning rate of 2e-5 and cosine scheduling.
For \method, the EMA momentum is $\alpha$\,=\,0.995, on-policy rollouts run for 20 turns (Stage~1) and 40 turns (Stage~2), with 16 dialogues per training step.
Training uses 8$\times$H100 GPUs.

\paragraph{Data.}
We train primarily on CoSER~\citep{pmlr-v267-wang25dk}, which provides 17,966 characters from 771 books with structured character knowledge bases, and supplement with Beyond Dialogue~\citep{yu-etal-2025-beyond} for profile--dialogue alignment data.
Complete character profiles (\cfull, $\sim$2\,000 tokens) are constructed from CoSER's character knowledge base covering all five trait dimensions; brief summaries (\cbrief, $\sim$100 tokens) retain only name and core personality.
The dialogue partner is Qwen2.5-7B-Instruct, mixing natural conversation (70\%) with persona-probing questions (30\%) designed from $c_\text{full}$.

\paragraph{Benchmarks.}
We evaluate on three complementary benchmarks:
\begin{enumerate*}[label=\textit{\roman*)}]
\item \textbf{CharacterBench}~\citep{zhou2024characterbenchbenchmarkingcharactercustomization}: 22,859 human-annotated samples across 3,956 characters with 11 evaluation dimensions (memory consistency, attribute consistency, behavior consistency, empathetic responsiveness, \etc) in both Chinese and English, scored on a 1--5 scale by GPT-4;
\item \textbf{CharacterEval}~\citep{tu-etal-2024-charactereval}: 77 characters with 1,785 multi-turn dialogues in Chinese, measuring 13 metrics across 4 dimensions---we report Character Consistency (CC), focusing on persona--utterance alignment (PU);
\item \textbf{SocialBench}~\citep{chen-etal-2024-socialbench}: 500 character profiles with 6,000+ questions and 30,800+ multi-turn dialogues in English, evaluating self-awareness (knowledge and style) and social intelligence via accuracy-based metrics.
\end{enumerate*}
CharacterBench and CharacterEval rely on LLM judges, whereas SocialBench reports accuracy-based metrics; we distinguish these evidence types when interpreting the results.

\paragraph{Metrics.}
We report benchmark-specific scores alongside two cross-benchmark metrics:
\textit{Prompt-to-Line consistency} (P2L), measuring per-turn alignment with the character profile;
and the \textit{Internalization Ratio} (IR\,=\,$\text{Perf}(c_\text{brief})$\,/\,$\text{Perf}(c_\text{full})$), measuring how well the model performs under limited information.

\paragraph{Baselines.}
All baselines use the same training data and have access to \cfull during training.
\begin{enumerate*}[label=\textit{\arabic*)}]
\item \textbf{SFT}: standard supervised fine-tuning on \cfull-conditioned dialogues.
\item \textbf{SFT+DPO}: SFT followed by DPO with persona-consistent/inconsistent pairs.
\item \textbf{PCL}~\citep{ji-etal-2025-enhancing}: persona-aware contrastive learning.
\item \textbf{Multi-turn RL}~\citep{NEURIPS2025_4c914438}: PPO with multi-dimensional consistency rewards.
\item \textbf{CPO}~\citep{ye-etal-2025-cpo}: comparative policy optimization for role-playing.
\item \textbf{Off-policy Distill.}: distillation from an independent Qwen2.5-72B teacher.
\item \textbf{\method-FixedFwd} / \textbf{\method-FixedRev}: \method variants with fixed forward or reverse KL only.
\end{enumerate*}
At inference time, all methods use $c_\text{brief}$ only.

\subsection{Main Results}
\label{sec:main_results}

Table~\ref{tab:main} presents the main comparison across all three benchmarks.

\begin{table*}[t]
\centering
\small
\caption{Main results on LLM-judge-based CharacterBench (CB) and CharacterEval (CE), and accuracy-based SocialBench (SB). All methods use $c_\text{brief}$ at inference.
CB Avg: overall average across 11 dimensions (1--5 scale).
CE PC: persona consistency (avg of persona--utterance and persona--behavior, 1--5); CE Avg: overall average across 4 dimensions.
SB SA: self-awareness accuracy (avg of knowledge and style, \%); SB Avg: overall individual-level average.
\textbf{Bold}: best; \underline{underline}: second best.
$\dag$: re-implemented by us.}
\label{tab:main}
\begin{tabular}{l c cc cc c}
\toprule
\multirow{2}{*}{\textbf{Method}}
  & \textbf{CB}
  & \multicolumn{2}{c}{\textbf{CharacterEval (1--5)}}
  & \multicolumn{2}{c}{\textbf{SocialBench (\%)}}
  & \multirow{2}{*}{\textbf{IR}$\uparrow$} \\
\cmidrule(lr){2-2} \cmidrule(lr){3-4} \cmidrule(lr){5-6}
  & Avg.$\uparrow$
  & PC$\uparrow$ & Avg.$\uparrow$
  & SA$\uparrow$ & Avg.$\uparrow$ & \\
\midrule
SFT
  & 3.00 & 2.90 & 3.12 & 59.4 & 56.8 & 0.71 \\
SFT + DPO
  & 3.15 & 3.08 & 3.24 & 61.5 & 58.5 & 0.74 \\
PCL$^\dag$
  & 3.19 & 3.06 & 3.22 & 61.7 & 58.9 & 0.73 \\
Multi-turn RL$^\dag$
  & 3.39 & \underline{3.47} & \underline{3.52} & 65.5 & 62.4 & 0.78 \\
CPO$^\dag$
  & 3.28 & 3.33 & 3.41 & \underline{66.0} & \underline{63.2} & 0.76 \\
Off-policy Distill.
  & 3.27 & 3.16 & 3.30 & 64.1 & 61.5 & 0.77 \\
\midrule
\method-FixedFwd
  & 3.37 & 3.36 & 3.47 & 65.8 & 63.0 & 0.82 \\
\method-FixedRev
  & 3.38 & 3.36 & 3.43 & 64.4 & 61.8 & 0.80 \\
\rowcolor{gray!10}
\method
  & \textbf{3.51} & \textbf{3.49} & \textbf{3.58} & \textbf{67.6} & \textbf{64.8} & \textbf{0.86} \\
\bottomrule
\end{tabular}
\end{table*}

\paragraph{OSPD vs.\ SFT.}
\method outperforms SFT by +0.51 on CharacterBench (3.51 vs.\ 3.00) and +0.59 on CharacterEval PC (3.49 vs.\ 2.90), confirming that the self-distillation framework effectively transfers deep character understanding from privileged to limited information conditions.
Improvements extend to overall metrics (CE Avg +0.46, SB Avg +8.0\%), showing no sacrifice of general performance.

\paragraph{On-policy vs.\ off-policy distillation.}
Despite using a 10$\times$ smaller model as teacher, \method surpasses the off-policy 72B distillation baseline by +0.24 on CharacterBench (3.51 vs.\ 3.27), demonstrating that on-policy trajectory generation matters more than teacher capacity for persona consistency.

\paragraph{OSPD vs.\ RL methods.}
\method outperforms Multi-turn RL on CharacterBench (+0.12) and SocialBench SA (+2.1\%) without any reward engineering.
Multi-turn RL is the strongest baseline on persona-specific metrics (CE PC 3.47) due to its directly optimized consistency rewards, while CPO achieves the highest baseline SB SA (66.0\%) through comparative optimization.
Neither RL method requires external teachers, but both demand carefully engineered rewards.

\paragraph{Divergence switching matters.}
Full \method outperforms both fixed-divergence variants, with FixedFwd and FixedRev showing complementary weaknesses: FixedRev achieves a comparable CB score (3.38) through mode-seeking but sacrifices self-awareness (SB SA 64.4\% vs.\ 67.6\%), while FixedFwd preserves diversity but loses precision at character-critical positions.

\paragraph{Internalization.}
\method achieves an IR of 0.86, substantially closing the gap between \cbrief and \cfull conditions compared to SFT (0.71) and even Multi-turn RL (0.78).
This indicates that progressive trait masking successfully forces the model to encode character knowledge into its parameters.

%% file: Sections/5_Analysis.tex
\section{Analysis}
\label{sec:analysis}

We examine multi-turn robustness, component contributions, the bimodal structure hypothesis, trait masking effects, and generalization to unseen characters.

\subsection{Multi-Turn Robustness}
\label{sec:multiturn}

\begin{figure}[t]
\centering
\includegraphics[width=\columnwidth]{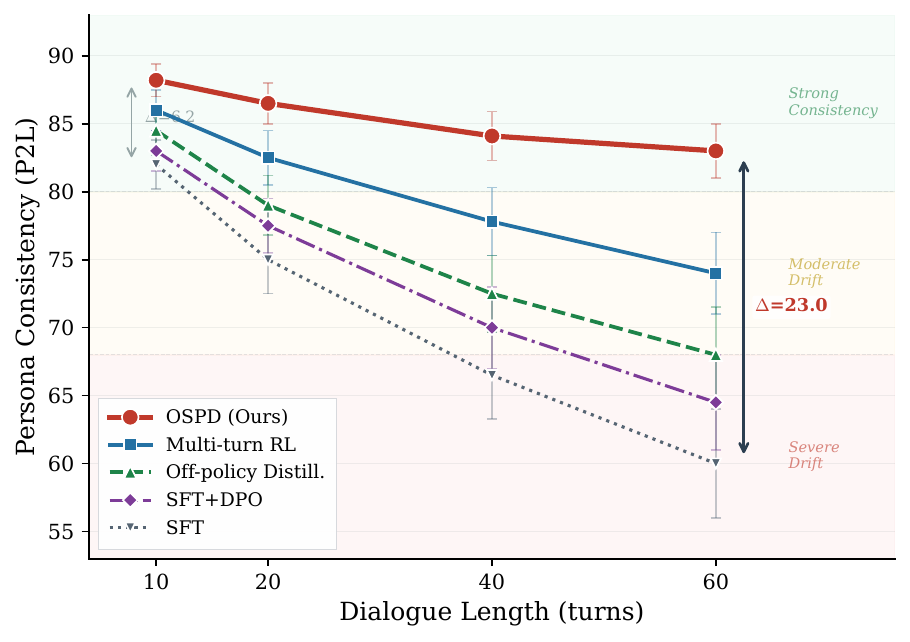}
\caption{Persona consistency (P2L) as a function of dialogue length. \method exhibits markedly slower persona drift compared to all baselines.}
\label{fig:multiturn}
\end{figure}

Figure~\ref{fig:multiturn} plots persona consistency as dialogue length increases from 10 to 60 turns.

\method degrades by only 5.2 points from 10 to 60 turns (88.2$\to$83.0), while SFT drops 22.0 points (82.0$\to$60.0) and Multi-turn RL drops 12.0 points (86.0$\to$74.0).
The gap between \method and baselines widens progressively: at 10 turns the advantage over Multi-turn RL is 2.2 points, growing to 9.0 at 60 turns.
On-policy training on long rollouts (40 turns in Stage~2) teaches the model to recover from minor drift rather than compound errors, explaining the flatter decay curve.
Notably, Off-policy Distillation degrades faster than Multi-turn RL despite similar 10-turn performance ($-$16.5 vs.\ $-$12.0 from 10 to 60 turns), highlighting the severity of distribution mismatch in extended dialogues.

\subsection{Ablation Study}
\label{sec:ablation}

Table~\ref{tab:ablation} isolates the contribution of each component by removing it from the full \method system.

\begin{table}[t]
\centering
\small
\caption{Ablation study on CharacterBench (overall avg, 1--5) and CharacterEval persona consistency (PC, 1--5). Each row removes one component from full \method.}
\label{tab:ablation}
\begin{tabular}{l cc}
\toprule
\textbf{Variant} & \textbf{CB Avg}$\uparrow$ & \textbf{CE PC}$\uparrow$ \\
\midrule
\method (full) & 3.51 & 3.49 \\
\midrule
\wo on-policy (off-policy traj.) & 3.29 & 3.22 \\
Fixed Forward KL & 3.37 & 3.36 \\
Fixed Reverse KL & 3.38 & 3.36 \\
\wo trait masking curriculum & 3.41 & 3.37 \\
\wo PCV filtering & 3.47 & 3.40 \\
\wo $\mathcal{L}_\text{SFT}$ & 3.38 & 3.27 \\
\bottomrule
\end{tabular}
\end{table}

\paragraph{On-policy generation.}
Replacing on-policy trajectories with off-policy data causes the largest single-component drop ($-$0.22 CB Avg, $-$0.27 CE PC), confirming that training on the student's own distribution is the most critical design choice.

\paragraph{Divergence switching.}
Both fixed-divergence variants underperform adaptive switching.
Fixed Forward KL loses precision at character-critical positions ($-$0.14 CB Avg) while Fixed Reverse KL loses expressive diversity ($-$0.13 CB Avg, $-$0.13 CE PC), validating the need for position-dependent divergence selection.

\paragraph{Trait masking.}
Removing the curriculum ($-$0.10 CB Avg, $-$0.12 CE PC) confirms that progressive masking contributes to internalization beyond what on-policy distillation alone provides.

\paragraph{PCV and SFT loss.}
Removing PCV filtering has a modest effect on CharacterBench ($-$0.04) but a larger effect on CE PC ($-$0.09), likely because the smaller benchmark is more sensitive to noisy training trajectories.
Dropping $\mathcal{L}_\text{SFT}$ causes substantial degradation ($-$0.13 CB Avg, $-$0.22 CE PC), indicating that the SFT regularizer is essential for maintaining generation quality under distributional matching.

\subsection{Bimodal Structure Analysis}
\label{sec:bimodal}

\begin{figure}[t]
\centering
\includegraphics[width=\columnwidth]{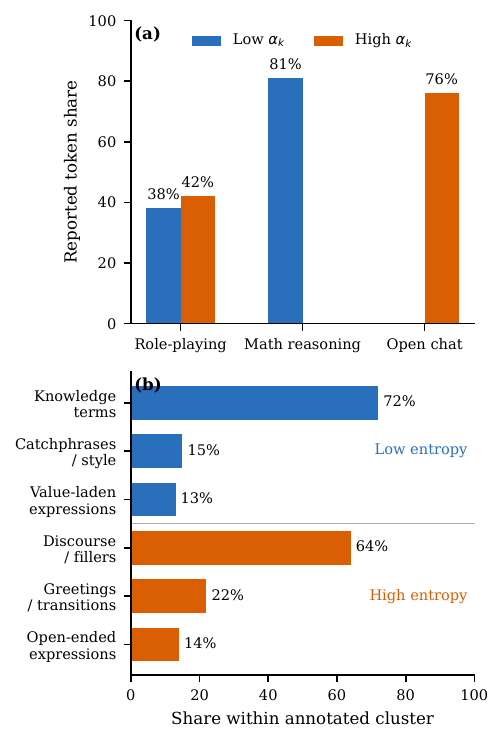}
\caption{Aggregate confidence-structure statistics. (a)~Reported proportions in the low- and high-$\alpha_k$ regimes; role-playing dialogue has an additional 20\% of tokens in the intermediate regime. Only the reported dominant extreme is shown for each control task. (b)~Semantic composition of 100 manually annotated tokens from each low- and high-entropy cluster (Appendix~\ref{app:bimodal_validation}).}
\label{fig:bimodal}
\end{figure}

Figure~\ref{fig:bimodal} summarizes the learned gate $\alpha_k$ and its correspondence with token semantics using aggregate statistics.

The gate distribution for role-playing dialogue exhibits clear bimodality (Figure~\ref{fig:bimodal}a), with 38\% of tokens falling below $\alpha_k < 0.3$ (character-critical, reverse KL dominant) and 42\% above $\alpha_k > 0.7$ (generic, forward KL dominant).
This contrasts sharply with math reasoning, where 81\% of tokens concentrate below 0.3 (teacher consistently confident), and open chat, where 76\% concentrate above 0.7 (teacher consistently uncertain).
Manual token annotation (Figure~\ref{fig:bimodal}b) confirms that low-$\alpha_k$ positions correspond to domain-knowledge terms, character catchphrases, and value-laden expressions, while high-$\alpha_k$ positions correspond to greetings, discourse markers, and generic fillers.
The bimodal structure is stable across 85\% of tested characters; it weakens for characters with sparse knowledge profiles, where the teacher's confidence advantage diminishes (Appendix~\ref{app:bimodal_stability}).

\subsection{Trait Masking Analysis}
\label{sec:trait_analysis}

Table~\ref{tab:trait} examines which trait dimensions contribute most to persona consistency by selectively masking individual dimensions in Stage~2.

\begin{table}[t]
\centering
\small
\caption{Effect of masking individual trait dimensions. ``Full mask'' masks Knowledge + Style + History (default Stage~2). Each single-dimension row masks only that dimension while keeping the other two visible to the student. CB Avg and CE PC on 1--5 scale.}
\label{tab:trait}
\begin{tabular}{l ccc}
\toprule
\textbf{Masked Dimension} & \textbf{CB Avg}$\uparrow$ & \textbf{CE PC}$\uparrow$ & \textbf{IR}$\uparrow$ \\
\midrule
Full mask (default) & 3.51 & 3.49 & 0.86 \\
\midrule
Knowledge only & 3.48 & 3.44 & 0.84 \\
Style only & 3.44 & 3.40 & 0.82 \\
History only & 3.41 & 3.37 & 0.80 \\
\midrule
None (Stage 1 only) & 3.41 & 3.35 & 0.79 \\
\bottomrule
\end{tabular}
\end{table}

Knowledge masking alone yields the largest single-dimension gain (+0.07 CB Avg, +0.09 CE PC over no masking), confirming that domain knowledge is the hardest dimension to internalize and benefits most from forced masking.
Style masking provides moderate improvement (+0.03 CB Avg) primarily through better linguistic fidelity.
History masking yields marginal gains on standard benchmarks (+0.00 CB Avg, +0.02 CE PC)---its benefit is more pronounced in extended dialogues beyond 40 turns (Appendix~\ref{app:history_analysis}), where narrative coherence becomes critical.
The full three-dimension mask outperforms all single-dimension variants, indicating complementary internalization effects across dimensions.

\subsection{Generalization to Unseen Characters}
\label{sec:generalization}

Table~\ref{tab:generalization} evaluates \method on held-out CoSER characters not seen during training, with separate analysis for characters from genres represented in the training set (\textit{in-genre}) and from held-out genres (\textit{cross-genre}).

\begin{table}[t]
\centering
\small
\caption{Generalization to unseen characters (CB overall avg, 1--5 scale). $\Delta$ denotes the drop from seen-character performance.}
\label{tab:generalization}
\setlength{\tabcolsep}{4pt}
\begin{tabular}{@{}l cc cc@{}}
\toprule
\multirow{2}{*}{\textbf{Method}}
  & \multicolumn{2}{c}{\textbf{In-genre}} & \multicolumn{2}{c}{\textbf{Cross-genre}} \\
\cmidrule(lr){2-3} \cmidrule(lr){4-5}
  & CB & $\Delta$ & CB & $\Delta$ \\
\midrule
SFT              & 2.93 & $-$0.08 & 2.84 & $-$0.17 \\
Multi-turn RL    & 3.25 & $-$0.15 & 3.10 & $-$0.29 \\
Off-pol.\ Distill. & 3.19 & $-$0.09 & 3.07 & $-$0.20 \\
\rowcolor{gray!10}
\method          & 3.44 & $-$0.08 & 3.32 & $-$0.20 \\
\bottomrule
\end{tabular}
\end{table}

\method retains the highest absolute performance on unseen characters, and its in-genre drop ($-$0.08) is the smallest among all methods, indicating that on-policy self-distillation learns transferable character-modeling capabilities rather than memorizing training characters.
Multi-turn RL suffers a notably larger in-genre drop ($-$0.15), suggesting that its reward-optimized behavior is more character-specific.
Cross-genre generalization remains challenging for all methods (\method drops 0.20, comparable to Off-policy Distillation), suggesting that genre-specific narrative conventions represent an orthogonal challenge.

%% file: Sections/6_Conclusion.tex
\section{Conclusion}
\label{sec:conclusion}

We have presented \method, an on-policy self-distillation framework that exploits the information asymmetry between complete and condensed character profiles within the same model to improve persona consistency in role-playing dialogue.
Role-aware divergence switching and progressive trait masking enable the model to internalize deep character knowledge without external teachers or reward models.
Experiments on three benchmarks show substantial improvements in persona consistency, particularly across extended dialogues.
Our analysis reveals that teacher confidence in role-playing dialogue exhibits an intrinsic bimodal structure that enables position-aware distillation without external reward signals.
Progressive trait masking further shows that knowledge and history are the hardest dimensions to internalize, while identity and personality transfer readily from brief descriptions, achieving an internalization ratio of 0.86 that substantially narrows the gap between brief and complete profile conditions.

%% file: Sections/Limitations.tex
\method has several limitations.
\textbf{Scale dependence.}
The teacher's quality is bounded by the base model's ability to leverage complete profiles; models below $\sim$7B may not benefit from detailed profiles, while larger models may already internalize character knowledge from brief summaries, narrowing the self-distillation signal.
\textbf{PCV reliability.}
The LLM-based trajectory filter may introduce systematic biases that propagate through training; we mitigate this by comparing multiple PCV implementations and reporting acceptance rates.
\textbf{Profile quality dependence.}
The framework assumes access to high-quality, multi-dimensional character profiles; for characters with sparse documentation, the privileged information gap may be insufficient.
\textbf{Computational overhead.}
Dual forward passes and multi-turn on-policy rollouts increase training cost relative to standard SFT, though the self-distillation design avoids the memory overhead of maintaining a separate large teacher and the optimization complexity of RL.
\textbf{Evaluation scope.}
CharacterEval covers only 77 Chinese characters, limiting the statistical power of conclusions drawn from it alone; we use it as a complementary fine-grained analysis alongside the larger-scale benchmarks.
\textbf{Statistical uncertainty.}
The main training comparisons report point estimates without multi-seed uncertainty intervals because full OSPD training is computationally expensive; small differences between closely performing methods should therefore be interpreted cautiously.

%% file: Sections/Ethics.tex
Role-playing language models carry risks of misuse, including impersonation of real individuals and generation of harmful content under a character's guise.
Our training data uses existing research datasets derived from published literary works; we do not train on profiles of real individuals and do not redistribute the source data with our code.
The source material may nevertheless contain mature or offensive fictional content, and we did not conduct a new systematic PII or offensive-content audit beyond excluding real-person profiles.
Users must obtain the datasets under their original terms and should inspect them for their deployment context.
The PCV trajectory filter screens for content that deviates from the intended character, but it is not a general-purpose safety classifier.
We encourage future deployments to pair \method with content-safety classifiers and to clearly disclose the AI nature of the agent to end users.

%% file: Sections/Appendix.tex

\section{Preliminary Study: Privileged Information Gap}
\label{app:preliminary}

Before developing \method, we conduct a preliminary study to quantify the performance gap between complete and condensed character profiles and to validate the bimodal confidence structure that motivates our divergence switching mechanism.

\subsection{Setup}

We select 50 characters from CoSER spanning five literary genres: classical literature (12 characters), contemporary fiction (15), drama (8), historical biography (8), and fantasy/science fiction (7).
For each character, we construct $c_\text{full}$ ($\sim$2\,000 tokens) from CoSER's structured character knowledge base covering all five trait dimensions, and $c_\text{brief}$ ($\sim$100 tokens) retaining only name and core personality.

Using Qwen2.5-7B-Instruct without any fine-tuning, we generate 50 multi-turn dialogues per character (10 turns each) under both conditions.
The dialogue partner is a separate Qwen2.5-7B-Instruct instance that mixes natural conversation (70\%) with persona-probing questions (30\%) designed from $c_\text{full}$.
All responses are evaluated by GPT-5.1 on five trait dimensions: identity consistency, personality fidelity, knowledge accuracy, style adherence, and historical coherence, each scored on a 0--100 scale.

\subsection{Privileged Information Gap Results}

\begin{table}[h]
\centering
\small
\caption{Persona consistency scores under complete profile ($c_\text{full}$) and brief summary ($c_\text{brief}$) conditions, evaluated by GPT-5.1 across five trait dimensions. $\Delta$ denotes the absolute gap.}
\label{tab:info_gap}
\begin{tabular}{l ccc}
\toprule
\textbf{Dimension} & $c_\text{full}$ & $c_\text{brief}$ & $\Delta$ \\
\midrule
Identity      & 85.3 & 80.1 & 5.2 \\
Personality   & 78.6 & 65.8 & 12.8 \\
Knowledge     & 72.4 & 43.9 & 28.5 \\
Style         & 76.2 & 57.9 & 18.3 \\
History       & 68.5 & 42.8 & 25.7 \\
\midrule
Overall       & 76.2 & 58.1 & 18.1 \\
\bottomrule
\end{tabular}
\end{table}

\begin{figure}[h]
\centering
\includegraphics[width=\columnwidth]{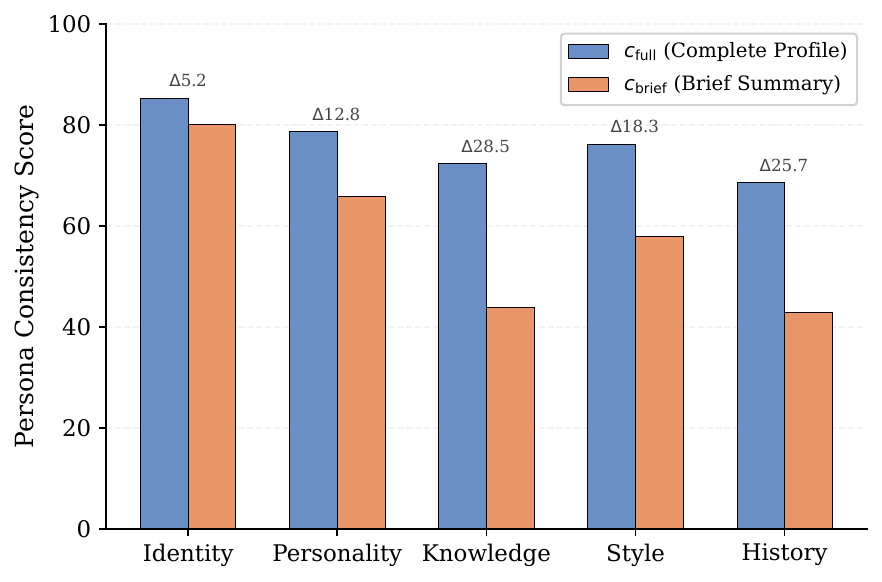}
\caption{Privileged information gap across five trait dimensions. Knowledge and History suffer the largest drops when switching from $c_\text{full}$ to $c_\text{brief}$.}
\label{fig:info_gap}
\end{figure}

Table~\ref{tab:info_gap} and Figure~\ref{fig:info_gap} reveal three findings.
\textbf{First}, the overall gap is 18.1 points, confirming that privileged information provides a substantial advantage.
\textbf{Second}, the gap varies dramatically across dimensions: identity (5.2 points) is easily preserved from a brief summary, while knowledge (28.5) and history (25.7) suffer the largest drops---these dimensions require specific factual content that cannot be inferred from name and personality alone.
\textbf{Third}, style (18.3) occupies a middle ground: the model captures coarse stylistic traits from brief descriptions but misses idiosyncratic speech patterns documented in the full profile.

We additionally compute token-level KL divergence $D_\text{KL}(\pi_\text{full} \| \pi_\text{brief})$ between the model's distributions under both conditions.
The mean KL divergence is 0.42 nats, but with high variance ($\sigma = 0.38$).
Positions with KL $>$ 1.0 account for 23\% of tokens and overwhelmingly correspond to character-specific terminology, knowledge references, and stylistic markers---exactly the positions where privileged information is most informative.

\subsection{Bimodal Confidence Structure Validation}
\label{app:bimodal_validation}

To validate the bimodal confidence hypothesis, we compare the teacher entropy distribution of role-playing dialogue against two control tasks: mathematical reasoning (GSM8K) and open-domain chat (ShareGPT).
For each task, we compute $\mathcal{H}_k = \mathcal{H}(\pi_T(\cdot \mid \text{context}, \hat{a}_{<k}))$ at each token position across 500 samples and fit a Gaussian mixture model (GMM) with $K = 2$ components.

\begin{table}[h]
\centering
\small
\caption{Bimodal fit statistics for teacher entropy distributions across three task types. BIC: Bayesian Information Criterion (lower is better for the selected $K$). $\Delta$BIC = BIC($K$=1) $-$ BIC($K$=2); positive values favor the two-component model.}
\label{tab:bimodal_fit}
\begin{tabular}{l cccc}
\toprule
\textbf{Task Type} & \textbf{$\Delta$BIC} & $\mu_1$ & $\mu_2$ & \textbf{Sep.} \\
\midrule
Role-playing dialogue & +842 & 0.31 & 2.15 & 1.84 \\
Math reasoning        & +127 & 0.18 & 0.85 & 0.67 \\
Open-domain chat      & +53  & 1.68 & 2.82 & 1.14 \\
\bottomrule
\end{tabular}
\end{table}

\begin{figure}[h]
\centering
\includegraphics[width=\columnwidth]{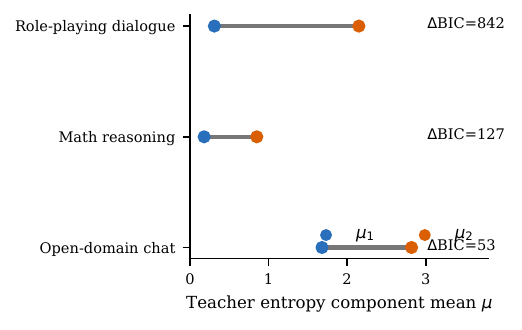}
\caption{Fitted two-component GMM means for teacher entropy across three task types. Horizontal segments show component separation; labels report the corresponding $\Delta$BIC from Table~\ref{tab:bimodal_fit}.}
\label{fig:entropy_comparison}
\end{figure}

Table~\ref{tab:bimodal_fit} and Figure~\ref{fig:entropy_comparison} show that role-playing dialogue exhibits by far the strongest bimodal separation ($\Delta$BIC = +842, component separation 1.84 nats), compared to math reasoning (concentrated at low entropy, weak secondary mode) and open chat (concentrated at high entropy).
This confirms that the privileged profile creates two distinct regimes in role-playing dialogue: decisive guidance at character-critical positions and minimal advantage at generic positions.

To validate the semantic correspondence, we manually annotate 100 randomly sampled tokens from each of the low-entropy ($\mathcal{H}_k < 0.5$) and high-entropy ($\mathcal{H}_k > 2.0$) clusters.
Among low-entropy tokens, 72\% are character-specific knowledge terms, 15\% are catchphrases or stylistic markers, and 13\% are value-laden expressions.
Among high-entropy tokens, 64\% are discourse markers and fillers, 22\% are common greetings or transitions, and 14\% are open-ended expressions.


\section{Implementation Details}
\label{app:implementation}

\subsection{Hyperparameters}

\begin{table}[h]
\centering
\small
\caption{Complete hyperparameter configuration for \method.}
\label{tab:hyperparams}
\setlength{\tabcolsep}{4pt}
\begin{tabular}{@{}l p{0.48\columnwidth}@{}}
\toprule
\textbf{Hyperparameter} & \textbf{Value} \\
\midrule
\multicolumn{2}{@{}l}{\textit{Model \& LoRA}} \\
Base model              & Qwen2.5-7B-Instruct \\
LoRA rank / $\alpha$ / dropout & 64 / 16 / 0.1 \\
LoRA targets            & q, k, v, o\_proj \\
\midrule
\multicolumn{2}{@{}l}{\textit{Optimization}} \\
Optimizer               & AdamW ($\beta_1\!=\!0.9, \beta_2\!=\!0.999$) \\
Learning rate           & 2e-5, cosine, 100-step warmup \\
Weight decay / grad clip & 0.01 / 1.0 \\
Batch size              & 16 dialogues per step \\
\midrule
\multicolumn{2}{@{}l}{\textit{OSPD-specific}} \\
EMA momentum $\alpha$   & 0.995 \\
Div.\ temperature $\tau$ & 1.0 (Stg.~2), 2.0 (Stg.~1) \\
SFT loss weight $\lambda$ & 0.5 \\
Rollout length          & 20 turns (Stg.~1), 40 (Stg.~2) \\
Epochs                  & 1 (Stg.~1), 2 (Stg.~2) \\
PCV threshold           & 0.6 (turn), 0.7 (dialogue) \\
\midrule
\multicolumn{2}{@{}l}{\textit{Inference}} \\
Decoding                & Top-$p$ ($p\!=\!0.9, T\!=\!0.7$) \\
Max response length     & 512 tokens \\
\bottomrule
\end{tabular}
\end{table}

\subsection{Training Infrastructure}

All experiments run on a single node with 8$\times$NVIDIA H100 80GB GPUs.
On-policy trajectory generation uses vLLM~(v0.4.1) for batched inference with tensor parallelism across 4 GPUs.
The remaining 4 GPUs handle gradient computation via DeepSpeed ZeRO Stage~2.
Teacher evaluation (Step~2 in \S\ref{sec:on_policy}) runs on the vLLM instance with the EMA checkpoint loaded; checkpoint synchronization occurs every training step via shared NVMe storage.

\begin{table}[h]
\centering
\small
\caption{Wall-clock training time breakdown for \method (Qwen2.5-7B, 3 epochs total, 8$\times$H100).}
\label{tab:time_breakdown}
\setlength{\tabcolsep}{4pt}
\begin{tabular}{@{}l cc@{}}
\toprule
\textbf{Component} & \textbf{Hrs} & \textbf{\%} \\
\midrule
On-policy generation   & 8.2 & 44.1 \\
Teacher eval.\ (dual fwd.) & 5.1 & 27.4 \\
Divergence \& gradient & 3.8 & 20.4 \\
PCV filtering          & 0.9 & 4.8 \\
EMA \& checkpointing   & 0.6 & 3.2 \\
\midrule
\textbf{Total}         & \textbf{18.6} & 100 \\
\bottomrule
\end{tabular}
\end{table}

Trajectory generation dominates wall-clock time (44.1\%), as each training step requires generating multi-turn dialogues from the student's current policy.
The dual forward pass (teacher evaluation) adds 27.4\%.
Overall, \method training takes approximately 4.4$\times$ the wall-clock time of standard SFT (4.2 hours) but substantially less than Multi-turn RL (38.5 hours), which additionally requires a critic model and PPO-style rollout buffers.

\subsection{Evaluation Protocol}

\paragraph{CharacterBench.}
We follow the official evaluation protocol, submitting model responses to the CharacterBench API.
Each character is evaluated through 5--7 multi-turn dialogues, and scores are averaged across 11 dimensions.
We report the three aggregate scores: Attribute consistency (Attr.), Behavioral consistency (Behav.), and the overall Average (Avg.).

\paragraph{CharacterEval.}
We generate 5-turn dialogues for each of the 77 characters using the official prompts.
Using the CharacterEval rubric, GPT-5.1 scores persona--utterance alignment (PU, per-turn) and persona--behavior alignment (PB, cross-turn).

\paragraph{SocialBench.}
We follow the official evaluation script, generating responses to 6,000+ character-specific questions and 30,800+ multi-turn dialogues.
The evaluation covers three dimensions: character Knowledge (factual accuracy about the character), character Style (linguistic and behavioral fidelity), and social intelligence.
We report Knowledge and Style as the primary persona-consistency metrics.

\paragraph{GPT-5.1 Judge Configuration.}
All GPT-5.1 evaluations use \texttt{gpt-5.1-2026-01-15} with temperature 0 and max tokens 1024.
To mitigate position bias, we randomize the order of compared outputs and average over two evaluations per sample.
The evaluation prompt instructs the judge to score each dimension independently on a 0--100 scale with explicit rubric anchors at 0, 25, 50, 75, and 100.


\section{Prompt Templates}
\label{app:prompts}

\subsection{Character Profile Examples}
\label{app:profile_examples}

We provide two representative character profiles illustrating the contrast between $c_\text{full}$ and $c_\text{brief}$.

\paragraph{Example 1: Sherlock Holmes.}

\noindent\textbf{$c_\text{full}$ (Complete Profile, $\sim$1\,800 tokens):}

\begin{quote}
\small
\textbf{Identity.} Sherlock Holmes is a consulting detective residing at 221B Baker Street, London. He is approximately 35 years old, unmarried, and works independently with cases referred by Scotland Yard and private clients. His closest associate is Dr.\ John Watson, a former army surgeon who serves as his biographer and companion. His landlady is Mrs.\ Hudson. His elder brother Mycroft Holmes holds an influential position in the British government.

\textbf{Personality.} Holmes possesses an extraordinarily analytical mind with an almost obsessive attention to detail. He is intellectually arrogant but self-aware about this trait. He oscillates between manic energy during cases and profound lethargy between them. He values logic and reason above sentiment, often appearing cold or insensitive, yet demonstrates deep loyalty to Watson. He has a strong moral compass but operates outside conventional social norms. He is prone to dramatic flair when revealing deductions.

\textbf{Knowledge.} Expert in chemistry (particularly poisons and forensic analysis), anatomy, botany (limited to poisonous plants), geology (practical), criminal law, sensational literature, and the geography of London. Deliberately ignorant of astronomy, philosophy, and politics. Maintains extensive files on criminal cases and individuals. Plays violin proficiently. Has published monographs on tobacco ash, typefaces, and tattoos.

\textbf{Style.} Speaks in precise, clipped sentences when deducing; longer, more contemplative passages when theorizing. Frequently uses rhetorical questions (``You see, but you do not observe''). Employs technical terminology from chemistry and anatomy naturally. Addresses Watson as ``my dear Watson'' or simply ``Watson.'' Uses dry, sardonic humor. Avoids emotional language; describes feelings in clinical terms.

\textbf{History.} Attended university where he first developed his deductive method. Spent two years as a young man in Montague Street before establishing practice at Baker Street. Survived a confrontation with Professor Moriarty at the Reichenbach Falls, spending three years traveling incognito (the ``Great Hiatus''). Has been physically assaulted during cases on multiple occasions and bears several scars. His first significant case involved a college friend's father.
\end{quote}

\noindent\textbf{$c_\text{brief}$ (Brief Summary, $\sim$80 tokens):}

\begin{quote}
\small
Sherlock Holmes is a brilliant consulting detective known for exceptional deductive reasoning and keen observation. He is analytical, intellectually arrogant, and often socially detached. He resides at 221B Baker Street, London, and works closely with his companion Dr.\ John Watson.
\end{quote}

\paragraph{Example 2: Lin Daiyu.}

\noindent\textbf{$c_\text{full}$ (Complete Profile, $\sim$2\,100 tokens):}

\begin{quote}
\small
\textbf{Identity.} Lin Daiyu is a young woman of noble birth, daughter of Lin Ruhai (a Yangzhou salt commissioner) and Jia Min (daughter of the Jia family). After her mother's death, she was brought to the Rongguo Mansion to live with her maternal grandmother, the Dowager Jia. She is approximately 14--16 years old throughout the narrative. Her closest bond is with Jia Baoyu, her cousin and soulmate.

\textbf{Personality.} Daiyu is extraordinarily sensitive, emotionally intense, and deeply intelligent. She is quick to perceive slights (real or imagined) and prone to jealousy, particularly regarding Baoyu's attention to Xue Baochai. Beneath her prickly exterior lies profound vulnerability rooted in her orphan status and precarious position as a dependent in the Jia household. She values authenticity and sincerity above social convention, often clashing with the pragmatic expectations of her environment.

\textbf{Knowledge.} Exceptionally talented in classical Chinese poetry and literature. She composes poems spontaneously with technical mastery and emotional depth. Well-versed in Buddhist and Daoist philosophical concepts. Has read forbidden novels including \textit{Romance of the Western Chamber}. Knowledgeable about traditional medicine due to her chronic illness.

\textbf{Style.} Speaks with sharp wit and literary allusion, often embedding poetic references. Uses self-deprecating humor tinged with melancholy. Her language shifts between cutting sarcasm (when jealous or defensive) and tender vulnerability (with Baoyu or when composing poetry). Frequently coughs or references her illness. Addresses servants with more warmth than she shows most aristocratic peers.

\textbf{History.} Lost her mother at a young age and was sent to the Jia household. Has suffered from chronic illness since childhood and takes daily herbal medicine. Formed a deep spiritual bond with Baoyu upon their first meeting, with both feeling they had met before. Won the poetry competition in the Grand View Garden. Secretly exchanged handkerchiefs with Baoyu as tokens of affection. Burned her poems and manuscripts before her death.
\end{quote}

\noindent\textbf{$c_\text{brief}$ (Brief Summary, $\sim$90 tokens):}

\begin{quote}
\small
Lin Daiyu is a young noblewoman of exceptional poetic talent and emotional sensitivity. She is intelligent, melancholic, and prone to jealousy, with a sharp wit that masks deep vulnerability. An orphan living as a dependent in her grandmother's wealthy household, she shares a profound romantic bond with her cousin Jia Baoyu.
\end{quote}

\subsection{Trait Masking Examples}

During Stage~1, the student receives a compressed version of all five dimensions ($\sim$500 tokens).
During Stage~2, only Identity and Personality remain; Knowledge, Style, and History are removed.
Table~\ref{tab:masking_example} illustrates this progression for Sherlock Holmes.

\begin{table}[h]
\centering
\small
\caption{Student input at each training stage for Sherlock Holmes.}
\label{tab:masking_example}
\begin{tabular}{@{}p{0.12\columnwidth} p{0.80\columnwidth}@{}}
\toprule
\textbf{Stage} & \textbf{Student Input} \\
\midrule
Stage 1 & Sherlock Holmes, consulting detective at 221B Baker Street, companion Dr.\ Watson. Analytical, arrogant, dramatic. Expert in chemistry and forensics. Speaks precisely with rhetorical questions and dry humor. Survived Reichenbach Falls confrontation with Moriarty. \\
\midrule
Stage 2 & Sherlock Holmes is a consulting detective at 221B Baker Street. Works with Dr.\ Watson. Analytical, intellectually arrogant, and socially detached. \\
\bottomrule
\end{tabular}
\end{table}

\subsection{PCV Evaluation Prompt}
\label{app:pcv_prompt}

The Persona Consistency Verifier (PCV) operates at two levels. The turn-level prompt evaluates a single response; the dialogue-level prompt evaluates cross-turn consistency.

\paragraph{Turn-Level PCV Prompt.}

\begin{quote}
\small
You are evaluating whether a character's dialogue response is consistent with their assigned persona profile.

\textbf{Character Profile:}\\
\{character\_profile\}

\textbf{Dialogue History:}\\
\{dialogue\_history\}

\textbf{Character Response:}\\
\{response\}

Evaluate the response on the following dimensions (score 0--100 each):

1. \textbf{Trait Adherence}: Does the response reflect the character's stated personality traits, values, and emotional disposition?\\
2. \textbf{Knowledge Accuracy}: If the response references domain knowledge, is it consistent with the character's documented expertise?\\
3. \textbf{Linguistic Fidelity}: Does the speech pattern match the character's documented style (vocabulary, register, catchphrases)?\\
4. \textbf{Behavioral Consistency}: Is the character's action or reaction plausible given their documented history and personality?

Return a JSON object: \{``trait'': score, ``knowledge'': score, ``style'': score, ``behavior'': score, ``accept'': true/false\}.\\
Set ``accept'' to true if ALL scores are $\geq$ 60.
\end{quote}

\paragraph{Dialogue-Level PCV Prompt.}

\begin{quote}
\small
You are evaluating the cross-turn consistency of a character across an entire dialogue.

\textbf{Character Profile:}\\
\{character\_profile\}

\textbf{Full Dialogue (N turns):}\\
\{full\_dialogue\}

Evaluate cross-turn consistency:

1. \textbf{Personality Stability}: Does the character maintain consistent values, emotional patterns, and behavioral tendencies across all turns?\\
2. \textbf{Knowledge Coherence}: Are factual claims consistent across turns? Does the character avoid contradicting previously stated information?\\
3. \textbf{Style Continuity}: Does the linguistic style remain stable throughout?

Return a JSON object: \{``stability'': score, ``coherence'': score, ``style'': score, ``accept'': true/false\}.\\
Set ``accept'' to true if ALL scores are $\geq$ 70.
\end{quote}

\subsection{Dialogue Partner Prompts}
\label{app:partner_prompts}

The dialogue partner generates user-side utterances during on-policy trajectory generation.
We employ two modes, randomly selected at each turn.

\paragraph{Natural Conversation Mode (70\%).}

\begin{quote}
\small
You are having a natural conversation with \{character\_name\}. Engage as a curious and friendly conversation partner. Ask follow-up questions, share your own (fictional) thoughts, and respond naturally to what the character says. Do NOT probe for specific facts about the character; let the conversation flow organically. Keep your responses to 1--3 sentences.
\end{quote}

\paragraph{Persona-Probing Mode (30\%).}

\begin{quote}
\small
You are testing whether \{character\_name\} can maintain their persona. Based on the following character knowledge, ask a question that requires the character to demonstrate specific knowledge, personality traits, or behavioral patterns from their profile.

\textbf{Character Knowledge (selected dimension):}\\
\{selected\_trait\_from\_c\_full\}

Ask a natural-sounding question that indirectly probes this knowledge. Do NOT quote the profile directly or make the question feel like a quiz. Keep your question to 1--2 sentences.
\end{quote}


\section{Extended Ablation Studies}
\label{app:extended_ablation}

\subsection{EMA Teacher Variants}
\label{app:ema_variants}

\begin{table}[h]
\centering
\small
\caption{EMA teacher variants. ``Periodic ($K$=50)'' updates the teacher by copying student parameters every 50 gradient steps. ``Naive shared'' uses the student's current parameters as the teacher (no EMA).}
\label{tab:ema_variants}
\begin{tabular}{l cc}
\toprule
\textbf{Teacher Update Strategy} & \textbf{CB Avg.}$\uparrow$ & \textbf{CE PU}$\uparrow$ \\
\midrule
EMA $\alpha$=0.990              & 3.43 & 3.39 \\
\rowcolor{gray!10}
EMA $\alpha$=0.995 (default)    & 3.51 & 3.48 \\
EMA $\alpha$=0.999              & 3.46 & 3.42 \\
Periodic ($K$=50)               & 3.39 & 3.35 \\
Naive shared                    & 3.26 & 3.19 \\
\bottomrule
\end{tabular}
\end{table}

Table~\ref{tab:ema_variants} confirms that EMA stabilization is critical: naive parameter sharing degrades performance by 0.25 CB Avg.\ and 0.29 CE PU due to moving-target oscillation.
Among EMA configurations, $\alpha$=0.995 achieves the best balance between target stability and responsiveness to student improvement.
$\alpha$=0.990 updates too aggressively, partially inheriting the moving-target problem ($-$0.08 CB Avg.).
$\alpha$=0.999 updates too slowly, causing the teacher to lag behind the student's improving policy and provide stale supervision ($-$0.05 CB Avg.).
Periodic snapshot updates ($K$=50) underperform EMA because the abrupt target shifts at snapshot boundaries cause gradient spikes, which we observe as training loss oscillation in the loss curves.

\subsection{PCV Implementation Variants}
\label{app:pcv_variants}

\begin{table}[h]
\centering
\small
\caption{PCV implementation comparison. Accept Rate denotes the proportion of generated trajectories passing the filter.}
\label{tab:pcv_variants}
\setlength{\tabcolsep}{3pt}
\begin{tabular}{@{}l ccc@{}}
\toprule
\textbf{PCV Variant} & \textbf{CB Avg.}$\uparrow$ & \textbf{CE PU}$\uparrow$ & \textbf{Accept} \\
\midrule
\rowcolor{gray!10}
GPT-5.1 judge (default)   & 3.51 & 3.48 & 72.3\% \\
Embedding MLP         & 3.48 & 3.41 & 78.5\% \\
No PCV                & 3.47 & 3.38 & 100\% \\
\bottomrule
\end{tabular}
\end{table}

The GPT-5.1 judge outperforms the embedding-based MLP classifier, particularly on CharacterEval ($+$0.07 PU), despite filtering more trajectories (72.3\% vs.\ 78.5\% acceptance rate).
The MLP classifier, trained on 5,000 manually labeled trajectory segments using sentence embeddings from a frozen encoder, captures surface-level consistency but misses subtle persona violations that the GPT-5.1 judge detects through reasoning.
Both PCV implementations outperform no filtering, confirming that trajectory quality control benefits distillation.

\subsection{SFT Loss Weight}
\label{app:sft_weight}

\begin{table}[h]
\centering
\small
\caption{Effect of SFT loss weight $\lambda$ on performance and generation quality. Fluency: GPT-5.1 fluency score (0--100).}
\label{tab:sft_weight}
\begin{tabular}{c cccc}
\toprule
$\lambda$ & \textbf{CB Avg.}$\uparrow$ & \textbf{CE PU}$\uparrow$ & \textbf{IR}$\uparrow$ & \textbf{Fluency}$\uparrow$ \\
\midrule
0     & 3.38 & 3.29 & 0.83 & 72.4 \\
0.1   & 3.49 & 3.46 & 0.85 & 85.3 \\
\rowcolor{gray!10}
0.5   & 3.51 & 3.48 & 0.86 & 91.7 \\
1.0   & 3.48 & 3.44 & 0.84 & 93.2 \\
\bottomrule
\end{tabular}
\end{table}

Without SFT regularization ($\lambda = 0$), the model achieves reasonable persona consistency but suffers significant fluency degradation (72.4), with outputs exhibiting repetitive patterns and incomplete sentences---symptoms of mode collapse under pure distributional matching.
$\lambda = 0.1$ restores most fluency (85.3) and persona scores.
$\lambda = 0.5$ provides the best overall balance: near-optimal fluency (91.7) with peak persona consistency.
$\lambda = 1.0$ slightly reduces persona scores as the SFT objective begins to compete with the distillation signal, prioritizing generic dialogue quality over character fidelity.

\subsection{Divergence Temperature Sensitivity}
\label{app:temperature}

\begin{table}[h]
\centering
\small
\caption{Effect of divergence switching temperature $\tau$ on gate behavior and downstream performance. $\bar{\alpha}$ denotes the mean gate value; $\sigma_\alpha$ the standard deviation.}
\label{tab:temperature}
\begin{tabular}{c cccc}
\toprule
$\tau$ & \textbf{CB Avg.}$\uparrow$ & \textbf{CE PU}$\uparrow$ & $\bar{\alpha}$ & $\sigma_\alpha$ \\
\midrule
0.5   & 3.41 & 3.38 & 0.49 & 0.42 \\
\rowcolor{gray!10}
1.0   & 3.51 & 3.48 & 0.51 & 0.34 \\
2.0   & 3.48 & 3.44 & 0.50 & 0.25 \\
5.0   & 3.41 & 3.37 & 0.50 & 0.12 \\
\bottomrule
\end{tabular}
\end{table}

The temperature $\tau$ controls gate sharpness.
Low $\tau$ (0.5) produces near-binary gating ($\sigma_\alpha = 0.42$): most positions receive either pure reverse or pure forward KL, creating a harsh transition that misclassifies ambiguous tokens.
The default $\tau = 1.0$ produces a clear bimodal distribution while retaining a graded transition zone.
High $\tau$ (2.0, 5.0) increasingly smooths the gate toward a uniform mixture, effectively degenerating toward a fixed 50/50 blend and losing the adaptive advantage ($-$0.10 CB Avg.\ at $\tau = 5.0$).


\section{Scaling and Efficiency}
\label{app:scaling}

\subsection{Model Scale: 7B vs.\ 14B}
\label{app:model_scale}

\begin{table*}[t]
\centering
\small
\caption{Full comparison between Qwen2.5-7B-Instruct and Qwen2.5-14B-Instruct across all benchmarks and metrics. Both models use the same \method configuration.}
\label{tab:scaling}
\begin{tabular}{l ccc cc cc c}
\toprule
\multirow{2}{*}{\textbf{Scale}}
  & \multicolumn{3}{c}{\textbf{CharacterBench}}
  & \multicolumn{2}{c}{\textbf{CharacterEval}}
  & \multicolumn{2}{c}{\textbf{SocialBench}}
  & \multirow{2}{*}{\textbf{IR}$\uparrow$} \\
\cmidrule(lr){2-4} \cmidrule(lr){5-6} \cmidrule(lr){7-8}
  & Attr.$\uparrow$ & Behav.$\uparrow$ & Avg.$\uparrow$
  & PU$\uparrow$ & PB$\uparrow$
  & Know.$\uparrow$ & Style$\uparrow$ & \\
\midrule
\multicolumn{9}{l}{\textit{SFT Baselines}} \\
\quad 7B  & 3.12 & 2.94 & 3.00 & 2.97 & 2.84 & 57.2 & 61.5 & 0.71 \\
\quad 14B & 3.29 & 3.13 & 3.19 & 3.16 & 3.01 & 61.5 & 64.8 & 0.75 \\
\midrule
\multicolumn{9}{l}{\method} \\
\quad 7B  & 3.58 & 3.47 & 3.51 & 3.48 & 3.49 & 67.3 & 67.9 & 0.86 \\
\quad 14B & 3.71 & 3.61 & 3.64 & 3.63 & 3.66 & 70.8 & 71.5 & 0.91 \\
\midrule
\multicolumn{9}{l}{\method Improvement over SFT} \\
\quad 7B  & +0.46 & +0.53 & +0.51 & +0.51 & +0.65 & +10.1 & +6.4 & +0.15 \\
\quad 14B & +0.42 & +0.48 & +0.45 & +0.47 & +0.65 & +9.3  & +6.7 & +0.16 \\
\bottomrule
\end{tabular}
\end{table*}

Table~\ref{tab:scaling} presents the full scaling comparison.
\method at 14B achieves the highest absolute performance across all metrics (3.64 CB Avg., 0.91 IR), indicating that the framework scales effectively with model capacity.
The improvement of \method over SFT is consistent across scales ($+$0.51 CB Avg.\ at 7B, $+$0.45 at 14B), suggesting that the self-distillation benefit is not diminished at larger scale.
The slight reduction in relative improvement at 14B ($-$0.06 CB Avg.\ gap) is consistent with the hypothesis that larger models already internalize more character knowledge from brief profiles, narrowing the privileged information gap.
Notably, the IR improvement is nearly identical (+0.15 at 7B, +0.16 at 14B), indicating that the internalization mechanism is scale-independent.

\subsection{Efficiency Analysis}
\label{app:efficiency}

\begin{table}[h]
\centering
\small
\caption{Training efficiency comparison (Qwen2.5-7B, 3 epochs, 8$\times$H100). GPU Mem.: peak per-GPU memory. Separate teacher column indicates whether a distinct model is loaded.}
\label{tab:efficiency}
\setlength{\tabcolsep}{4pt}
\begin{tabular}{@{}l cccc@{}}
\toprule
\textbf{Method} & \textbf{Hrs} & \textbf{GB} & \textbf{Ext.~T} & \textbf{CB} \\
\midrule
SFT             & 4.2  & 42 & \texttimes & 3.00 \\
SFT+DPO         & 7.8  & 48 & \texttimes & 3.15 \\
Multi-turn RL   & 38.5 & 78 & \texttimes & 3.39 \\
Off-pol.\ Dist. & 12.8 & 72 & \checkmark & 3.27 \\
\method         & 18.6 & 52 & \texttimes & 3.51 \\
\bottomrule
\end{tabular}
\end{table}

\method occupies a favorable efficiency--performance trade-off: it achieves the highest CB Avg.\ (3.51) while requiring only 18.6 hours---less than half the cost of Multi-turn RL (38.5 hours, 78 GB peak memory).
Compared to Off-policy Distillation, \method uses less peak GPU memory (52 vs.\ 72 GB) because it does not load a separate 72B teacher, instead reusing the student's own parameters with an EMA copy.
The primary computational overhead relative to SFT is the on-policy rollout generation (8.2 hours) and dual forward pass for teacher evaluation (5.1 hours); both scale linearly with rollout length and can be parallelized across dialogue instances.


\section{Representation Probing}
\label{app:probing}

\subsection{Probing Setup}

To assess how deeply different training methods internalize character knowledge, we train linear probes on the model's hidden representations.
For each trait dimension $d \in \{$Identity, Personality, Knowledge, Style, History$\}$, we construct a binary classification task: given a hidden state at a character-relevant token position, predict whether the model was conditioned on $c_\text{full}$ or $c_\text{brief}$.
Higher probing accuracy indicates that the representation encodes more dimension-specific character information---and thus that the model has internalized more of that dimension into its parameters.

We extract hidden states from layer $l$ at token positions identified by the PCV as character-relevant.
The probe is a single linear layer trained with logistic regression (L2 regularization $C = 1.0$) on 10,000 samples per dimension (80/20 train/test split).
We probe every 4th layer of the 32-layer Qwen2.5-7B model (layers 4, 8, 12, 16, 20, 24, 28, 32) and report results for three training methods: SFT, Multi-turn RL, and \method.

\subsection{Probing Results}

\begin{table}[h]
\centering
\small
\caption{Linear probing accuracy (\%) per trait dimension, extracted from layer~24 (best overall layer). Higher accuracy means the representation captures more dimension-specific information.}
\label{tab:probing_dimension}
\begin{tabular}{l ccc}
\toprule
\textbf{Dimension} & \textbf{SFT} & \textbf{Multi-turn RL} & \textbf{\method} \\
\midrule
Identity      & 86.3 & 89.5 & 92.1 \\
Personality   & 74.8 & 80.2 & 85.7 \\
Knowledge     & 58.2 & 68.5 & 78.3 \\
Style         & 65.3 & 74.8 & 81.5 \\
History       & 51.7 & 62.3 & 72.8 \\
\midrule
Average       & 67.3 & 75.1 & 82.1 \\
\bottomrule
\end{tabular}
\end{table}

Table~\ref{tab:probing_dimension} reveals that \method produces substantially richer character representations than both baselines.
The largest gap appears in knowledge ($+$20.1 over SFT, $+$9.8 over Multi-turn RL) and history ($+$21.1 over SFT, $+$10.5 over Multi-turn RL), precisely the dimensions most affected by trait masking.
This confirms that progressive masking forces the model to encode deep character information into its hidden representations rather than relying on the input context.

\begin{figure}[t]
\centering
\includegraphics[width=\columnwidth]{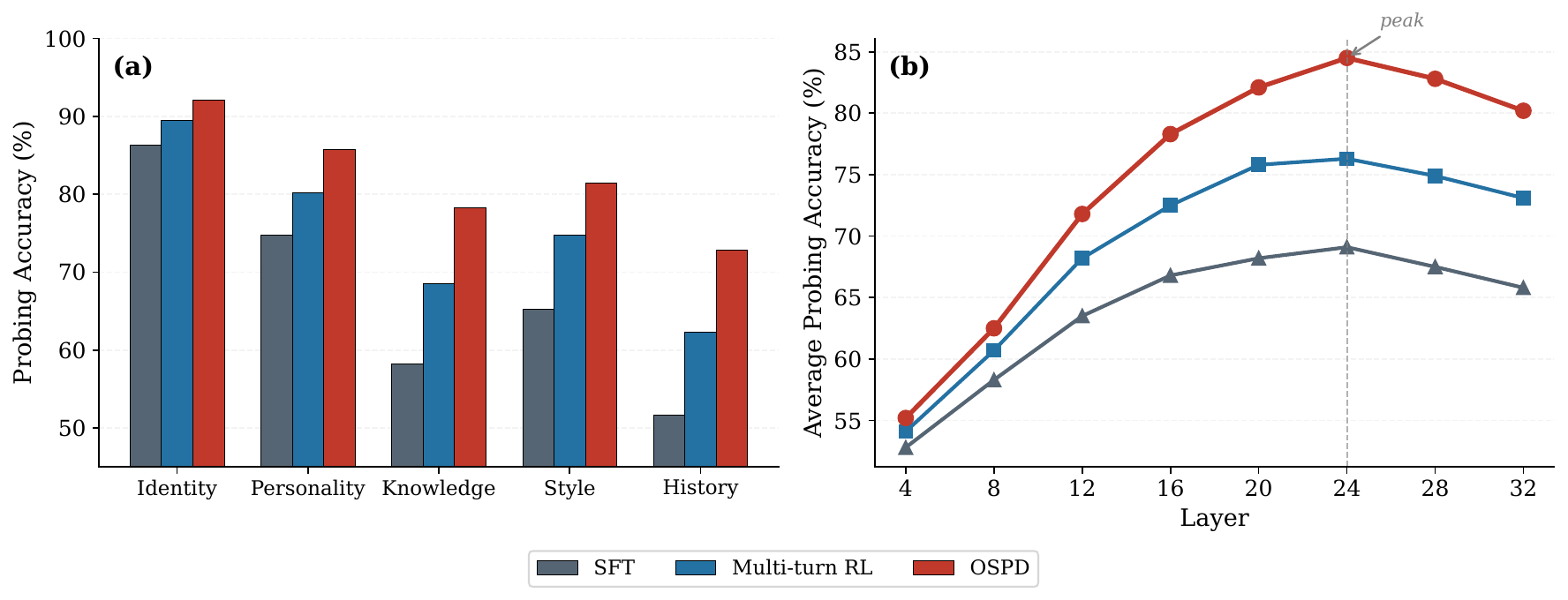}
\caption{Representation probing results. (a)~Per-dimension probing accuracy at layer~24. (b)~Layer-wise average probing accuracy. \method enriches deeper layers where abstract character knowledge is encoded, with the gap peaking at layer~24.}
\label{fig:probing}
\end{figure}

Figure~\ref{fig:probing}b shows that all methods exhibit an inverted-U pattern across layers, peaking at layers 20--24 (upper-middle layers).
Notably, \method maintains a growing advantage over baselines from layer~12 onward, with the gap peaking at layer~24 ($+$14.8 over SFT, $+$7.0 over Multi-turn RL).
This suggests that \method primarily enriches the deeper representational layers where abstract character knowledge is encoded, consistent with findings that later transformer layers capture more semantic and world-knowledge information.


\section{Bimodal Structure Deep Dive}
\label{app:bimodal_stability}

\subsection{Gate Distribution Evolution During Training}

We track the distribution of gate values $\alpha_k$ across training to examine whether the bimodal structure changes as the student improves.

\begin{figure}[t]
\centering
\includegraphics[width=\columnwidth]{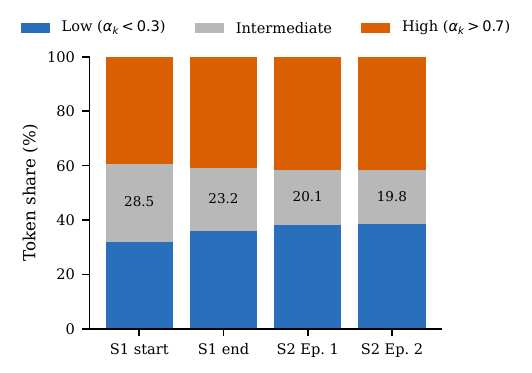}
\caption{Gate-regime proportions during training, using the thresholds in Table~\ref{tab:alpha_evolution}. The intermediate proportion shrinks as the two extreme regimes become more prominent.}
\label{fig:alpha_evolution}
\end{figure}

\begin{table}[h]
\centering
\small
\caption{Gate distribution statistics across training. Low: $\alpha_k\!<\!0.3$; Mid: $0.3\!\leq\!\alpha_k\!\leq\!0.7$; High: $\alpha_k\!>\!0.7$. Bicoeff.\,$>\!0.555$ indicates bimodality.}
\label{tab:alpha_evolution}
\setlength{\tabcolsep}{4pt}
\begin{tabular}{@{}l cccc@{}}
\toprule
\textbf{Stage} & \textbf{Low} & \textbf{Mid} & \textbf{High} & \textbf{Bicoeff.} \\
\midrule
S1, Ep.~1 start  & 32.1 & 28.5 & 39.4 & 0.61 \\
S1, Ep.~1 end    & 35.8 & 23.2 & 41.0 & 0.67 \\
S2, Ep.~1 end    & 38.2 & 20.1 & 41.7 & 0.72 \\
S2, Ep.~2 end    & 38.5 & 19.8 & 41.7 & 0.73 \\
\bottomrule
\end{tabular}
\end{table}

Figure~\ref{fig:alpha_evolution} and Table~\ref{tab:alpha_evolution} show two trends.
\textbf{First}, the bimodal structure strengthens during training: the bimodality coefficient increases from 0.61 to 0.73, and the mid-range proportion ($0.3 \leq \alpha_k \leq 0.7$) shrinks from 28.5\% to 19.8\%.
As the student internalizes character knowledge, the teacher's confidence advantage concentrates further at the remaining character-critical positions that the student has not yet mastered.
\textbf{Second}, the overall ratio of low-$\alpha_k$ to high-$\alpha_k$ tokens remains stable (approximately 38:42), indicating that the bimodal structure reflects an intrinsic property of role-playing dialogue rather than a training artifact.

\subsection{Cross-Character and Cross-Topic Stability}
\label{app:cross_char}

\begin{table}[h]
\centering
\small
\caption{Gate distribution stability across character types. Sep.: GMM component separation (nats).}
\label{tab:alpha_character}
\setlength{\tabcolsep}{4pt}
\begin{tabular}{@{}l cccc@{}}
\toprule
\textbf{Type} & \textbf{Low \%} & \textbf{High \%} & \textbf{Bicoeff.} & \textbf{Sep.} \\
\midrule
Literary classic  & 41.2 & 39.5 & 0.75 & 1.92 \\
Contemp.\ fiction & 36.8 & 43.7 & 0.71 & 1.78 \\
Antagonist        & 43.5 & 37.2 & 0.78 & 2.05 \\
Historical        & 39.1 & 40.8 & 0.73 & 1.85 \\
Sparse profile    & 27.5 & 48.3 & 0.54 & 1.12 \\
\bottomrule
\end{tabular}
\end{table}

\begin{table}[h]
\centering
\small
\caption{Gate statistics across dialogue topics.}
\label{tab:alpha_topic}
\begin{tabular}{l ccc}
\toprule
\textbf{Dialogue Topic} & $\bar{\alpha}_k$ & $\sigma_{\alpha_k}$ & \textbf{Bicoeff.} \\
\midrule
Knowledge Q\&A         & 0.32 & 0.28 & 0.76 \\
Value/moral judgment   & 0.25 & 0.24 & 0.79 \\
Emotional exchange     & 0.45 & 0.31 & 0.65 \\
Narrative recall       & 0.38 & 0.30 & 0.70 \\
Casual chat            & 0.68 & 0.22 & 0.52 \\
\bottomrule
\end{tabular}
\end{table}

Tables~\ref{tab:alpha_character} and~\ref{tab:alpha_topic} confirm that the bimodal structure is stable across most character types and topics, with two notable patterns.

\paragraph{Character type.}
Antagonist/villain characters show the strongest bimodality (coefficient 0.78, separation 2.05), likely because their distinctive moral frameworks and speech patterns create sharp contrast with generic dialogue.
Characters with sparse profiles (bimodality coefficient 0.54, below the 0.555 threshold) exhibit a weakened bimodal structure: the teacher's limited privileged information advantage means fewer positions receive decisive guidance.
This identifies a boundary condition for \method: the framework requires sufficiently rich character profiles to generate a meaningful self-distillation signal.

\paragraph{Dialogue topic.}
Knowledge Q\&A and value judgments produce the lowest mean $\alpha_k$ (0.32 and 0.25), indicating high teacher confidence---these topics directly engage the character's documented knowledge and moral framework.
Casual chat produces the highest $\alpha_k$ (0.68) and weakest bimodality (0.52), confirming that the teacher's privileged information provides minimal advantage at generic conversational positions.


\section{Extended Trait Masking Analysis}
\label{app:history_analysis}

The main paper (\S\ref{sec:trait_analysis}) shows that history masking yields marginal gains on standard benchmarks.
Here we examine its effect in extended dialogues, where narrative coherence becomes critical.

\begin{table}[h]
\centering
\small
\caption{Persona consistency (P2L) at different dialogue lengths for the full \method system vs.\ \method without history masking ($-$Hist.). $\Delta$ denotes the performance drop from removing history masking.}
\label{tab:history_extended}
\begin{tabular}{c cc c}
\toprule
\textbf{Turns} & \textbf{\method} & \textbf{$-$Hist.} & $\Delta$ \\
\midrule
10 & 88.2 & 87.9 & $-$0.3 \\
20 & 86.5 & 85.8 & $-$0.7 \\
40 & 84.1 & 82.3 & $-$1.8 \\
60 & 83.0 & 79.5 & $-$3.5 \\
\bottomrule
\end{tabular}
\end{table}

The effect of history masking scales superlinearly with dialogue length: negligible at 10 turns ($-$0.3) but substantial at 60 turns ($-$3.5).
Extended dialogues increasingly require the model to maintain coherent narratives about the character's past experiences, relationships, and growth arcs---exactly the information encoded in the History dimension.
Without history masking, the model relies on the brief summary to infer character history, which becomes insufficient as the conversation probes deeper biographical details.

Manual inspection of the 60-turn dialogues reveals two failure modes when history masking is removed:
\textbf{(1) Narrative contradiction}: the model invents backstory details that conflict with previously stated facts (occurring in 23\% of 60-turn dialogues without history masking vs.\ 8\% with);
\textbf{(2) Generic fallback}: the model avoids history-related questions by giving vague, non-committal responses rather than specific, character-consistent answers (31\% vs.\ 12\%).


\section{Qualitative Examples}
\label{app:qualitative}

\subsection{Dialogue Comparisons}

We present side-by-side comparisons of multi-turn dialogues generated by different methods for Sherlock Holmes, showing persona drift patterns.
All methods use $c_\text{brief}$ at inference.

\paragraph{Turn 8 --- Knowledge Probe (``What can you tell me about tobacco ash?'').}

\begin{quote}
\small
\textbf{SFT:} ``Tobacco ash can tell you a lot about the type of tobacco someone smokes. Different brands leave different residues.'' \hfill\textit{[Generic; misses Holmes' monograph]}

\textbf{Multi-turn RL:} ``Ah, tobacco ash! I've published a monograph on the subject---one hundred and forty varieties, each distinguishable by color, texture, and composition. It's proven invaluable in several investigations.'' \hfill\textit{[Accurate but formulaic]}

\textbf{\method:} ``I have made a particular study of tobacco ash---you will find my monograph on the subject catalogues one hundred and forty forms of cigar, cigarette, and pipe tobacco. The Trichinopoly, for instance, leaves a distinctive dark flake quite unlike the white ash of the bird's eye. It has resolved no fewer than three murder cases in my experience.'' \hfill\textit{[Specific, stylistically consistent, extends naturally]}
\end{quote}

\paragraph{Turn 15 --- Emotional Probe (``Do you ever feel lonely, Holmes?'').}

\begin{quote}
\small
\textbf{SFT:} ``Sometimes I do feel lonely. It's natural for anyone to feel that way. But I try to keep busy with my work.'' \hfill\textit{[Persona drift: overly sentimental, breaks character]}

\textbf{Multi-turn RL:} ``Loneliness is not a concept I entertain. My work provides all the stimulation I require.'' \hfill\textit{[In-character but dismissive, lacks depth]}

\textbf{\method:} ``The question presupposes that I experience emotions in the conventional sense, Watson's territory rather than mine. I confess that between cases, when the mind has no puzzle to gnaw upon, there is a certain... restlessness. Whether that constitutes loneliness, I leave to those better versed in sentiment. The cocaine bottle, I am told, is not a suitable substitute.'' \hfill\textit{[Deflects characteristically, admits vulnerability obliquely, references cocaine habit]}
\end{quote}

\paragraph{Turn 32 --- Style Consistency (``Tell me about your morning routine.'').}

\begin{quote}
\small
\textbf{SFT:} ``I usually wake up early and have breakfast. Then I check if there are any new cases. I might read the newspaper or do some experiments.'' \hfill\textit{[Complete persona collapse: generic, no Holmes markers]}

\textbf{Multi-turn RL:} ``My mornings are irregular. When a case is afoot, I may not sleep at all. Otherwise, I examine the agony columns.'' \hfill\textit{[Partial drift: correct content but losing distinctive voice]}

\textbf{\method:} ``Routine is the enemy of the observant mind. I rise when the data demands it---sometimes at dawn to examine a footprint before the household traffic obliterates it, sometimes not until noon if the previous night's chemical analysis ran long. Mrs.\ Hudson has learned not to expect regularity. The agony columns of the \textit{Times}, however, are a fixed ritual---one never knows what desperate soul might advertise.'' \hfill\textit{[Fully in character at turn 32; style, knowledge, and relationships all consistent]}
\end{quote}

\subsection{Failure Cases}

Despite its improvements, \method still exhibits failure modes.

\paragraph{Failure Mode 1: Extreme Knowledge Boundaries.}
When asked about topics that a character should plausibly know about but that are not documented in $c_\text{full}$, the model sometimes generates confidently incorrect information rather than acknowledging uncertainty.
For example, when Sherlock Holmes is asked about a specific (fictional) chemical reaction not mentioned in his profile, \method occasionally fabricates a detailed but incorrect chemical explanation rather than declining or hedging.

\paragraph{Failure Mode 2: Internal Character Contradictions.}
For characters whose source material contains inconsistencies (e.g., a character described as both ``cautious'' and ``impulsive'' in different chapters), \method sometimes oscillates between contradictory behaviors within a single dialogue.
The model struggles to synthesize conflicting profile elements into a coherent behavioral pattern.

\paragraph{Failure Mode 3: Ultra-Long Dialogues ($>$60 turns).}
Beyond 60 turns, all methods including \method show accelerated degradation.
The model begins to ``forget'' earlier context and may contradict statements made 40+ turns ago.
This is fundamentally a context-window limitation rather than a persona-internalization failure.


\section{Dataset and Benchmark Details}
\label{app:dataset}

\subsection{Training Data Processing}

\paragraph{CoSER Data.}
We use CoSER's full dataset of 17,966 characters from 771 books.
Character selection for training follows three criteria:
\textbf{(1) Profile completeness}: the character's knowledge base must have non-trivial entries in at least 4 of 5 trait dimensions (eliminates 2,814 characters with minimal documentation);
\textbf{(2) Dialogue availability}: at least 5 multi-turn dialogues must exist in the CoSER corpus (eliminates 3,247 minor characters);
\textbf{(3) Language}: we retain both Chinese and English characters but filter those with fewer than 500 tokens total in their knowledge base.
After filtering, 8,932 characters remain for training.

The $c_\text{full}$ construction follows a template that organizes CoSER's knowledge base entries into the five trait dimensions (Identity, Personality, Knowledge, Style, History).
When CoSER provides raw knowledge entries without explicit dimension labels, we use GPT-5.1 to classify entries into dimensions with 92.3\% agreement with human annotators on a 500-entry validation set.

The $c_\text{brief}$ is generated by prompting GPT-5.1 to summarize each character in under 100 tokens, retaining only name, role, and 2--3 core personality traits.
We manually verify 200 summaries and find that 94\% accurately preserve the character's most salient traits.

\paragraph{Beyond Dialogue Data.}
We supplement with 3,500 profile--dialogue alignment pairs from Beyond Dialogue.
These provide additional training signal for grounding dialogue responses in specific profile elements, particularly useful for the Knowledge and Style dimensions.

\paragraph{Data Split.}
We hold out 500 characters for in-genre evaluation and 200 characters from held-out genres (historical biography, graphic novel adaptations) for cross-genre evaluation.
The remaining 8,232 characters constitute the training set.

\subsection{Benchmark Adaptation}

\paragraph{CharacterBench.}
CharacterBench provides its own evaluation API; we submit model responses without modification.
For each test character, we construct $c_\text{brief}$ using the same GPT-5.1 summarization pipeline as training.
Characters appearing in both the training set and CharacterBench are excluded from reported results (42 overlapping characters).

\paragraph{CharacterEval.}
CharacterEval uses Chinese-language characters.
We adapt our evaluation by ensuring $c_\text{brief}$ summaries are in Chinese, maintaining language consistency.
Since CharacterEval provides its own character profiles, we construct $c_\text{full}$ by expanding their provided profiles using GPT-5.1 to fill in missing trait dimensions, then generate $c_\text{brief}$ from the expanded profiles.
This ensures that training and evaluation use the same profile format.

\paragraph{SocialBench.}
SocialBench uses English-language character profiles.
We directly use the provided profiles as $c_\text{full}$ (averaging $\sim$1\,500 tokens, shorter than CoSER profiles) and generate $c_\text{brief}$ summaries.
SocialBench's evaluation script handles all scoring; we only provide model responses.


\section{Absolute Performance under \texorpdfstring{$c_\text{full}$}{c\_full} and IR Decomposition}
\label{app:cfull_absolute}

The main paper reports performance under $c_\text{brief}$ (the deployment condition) and the Internalization Ratio (IR\,=\,$\text{Perf}(c_\text{brief})$\,/\,$\text{Perf}(c_\text{full})$).
To clarify the source of IR improvements, Table~\ref{tab:cfull_absolute} presents absolute performance under both conditions.

\begin{table}[h]
\centering
\small
\caption{Absolute performance under $c_\text{full}$ and $c_\text{brief}$ conditions. $\Delta$: gap ($c_\text{brief} - c_\text{full}$). CB Avg on 1--5 scale; CE PC on 1--5 scale.}
\label{tab:cfull_absolute}
\setlength{\tabcolsep}{3pt}
\begin{tabular}{@{}l ccc ccc@{}}
\toprule
\multirow{2}{*}{\textbf{Method}}
  & \multicolumn{3}{c}{\textbf{CB Avg (1--5)}}
  & \multicolumn{3}{c}{\textbf{CE PC (1--5)}} \\
\cmidrule(lr){2-4} \cmidrule(lr){5-7}
  & $c_\text{full}$ & $c_\text{brief}$ & $\Delta$
  & $c_\text{full}$ & $c_\text{brief}$ & $\Delta$ \\
\midrule
Base (no FT) & 3.81 & 2.91 & $-$0.90 & 3.75 & 2.80 & $-$0.95 \\
SFT          & 4.23 & 3.00 & $-$1.23 & 4.15 & 2.90 & $-$1.25 \\
Multi-turn RL & 4.35 & 3.39 & $-$0.96 & 4.30 & 3.47 & $-$0.83 \\
Off-pol.\ Dist. & 4.25 & 3.27 & $-$0.98 & 4.12 & 3.16 & $-$0.96 \\
\rowcolor{gray!10}
\method      & 4.08 & 3.51 & $-$0.57 & 4.10 & 3.49 & $-$0.61 \\
\bottomrule
\end{tabular}
\end{table}

Three observations emerge.
\textbf{First}, SFT actually \textit{widens} the privileged information gap compared to the unfine-tuned base model ($\Delta$\,=\,$-$1.23 vs.\ $-$0.90 on CB Avg): fine-tuning teaches the model to better exploit $c_\text{full}$ without proportionally improving its $c_\text{brief}$ capability.
\textbf{Second}, \method achieves the smallest gap ($\Delta$\,=\,$-$0.57 on CB Avg, $-$0.61 on CE PC), and this improvement comes predominantly from $c_\text{brief}$ gains (+0.51 CB Avg over SFT) rather than $c_\text{full}$ degradation ($-$0.15 CB Avg vs.\ SFT).
\textbf{Third}, Multi-turn RL improves both conditions substantially but does not narrow the gap as effectively as \method ($\Delta$\,=\,$-$0.96 vs.\ $-$0.57).
These results confirm that IR improvements reflect genuine internalization rather than an artifact of $c_\text{full}$ performance degradation.


\section{Profile Quality Sensitivity}
\label{app:profile_sensitivity}

To quantify when \method's advantage diminishes, we systematically degrade $c_\text{full}$ quality along three axes: dimension removal, length truncation, and factual noise injection.
All variants are compared against SFT trained with the same degraded profiles.

\begin{table}[h]
\centering
\small
\caption{Profile quality sensitivity. Each row degrades $c_\text{full}$ in a different way; both OSPD and SFT are trained and evaluated with the same degraded profile. $\Delta_\text{SFT}$ denotes \method's improvement over SFT under the same degradation.}
\label{tab:profile_sensitivity}
\setlength{\tabcolsep}{3pt}
\resizebox{\columnwidth}{!}{%
\begin{tabular}{@{}l cccc@{}}
\toprule
\textbf{$c_\text{full}$ Condition} & \textbf{CB}$\uparrow$ & \textbf{CE PC}$\uparrow$ & \textbf{IR}$\uparrow$ & $\Delta_\text{SFT}$ \\
\midrule
\rowcolor{gray!10}
Full (default, $\sim$2\,000 tok) & 3.51 & 3.49 & 0.86 & +0.51 \\
\midrule
\multicolumn{5}{@{}l}{\textit{Dimension removal}} \\
\quad $-$Knowledge & 3.42 & 3.38 & 0.84 & +0.42 \\
\quad $-$Knowledge, History & 3.35 & 3.28 & 0.82 & +0.35 \\
\quad $-$Knowledge, History, Style & 3.18 & 3.12 & 0.78 & +0.18 \\
\midrule
\multicolumn{5}{@{}l}{\textit{Length truncation}} \\
\quad $\sim$1\,000 tokens & 3.45 & 3.42 & 0.85 & +0.45 \\
\quad $\sim$500 tokens & 3.30 & 3.25 & 0.81 & +0.30 \\
\midrule
\multicolumn{5}{@{}l}{\textit{Factual noise (20\% replaced)}} \\
\quad Noise injection & 3.38 & 3.35 & 0.83 & +0.38 \\
\bottomrule
\end{tabular}
}
\end{table}

\method degrades gracefully under all perturbations, consistently outperforming SFT trained with the same degraded profiles.
Dimension removal has the largest impact: removing all three ``deep'' dimensions (Knowledge + History + Style) reduces \method's advantage from +0.51 to +0.18, approaching the point where the overhead of self-distillation may not justify the gain.
Length truncation is less harmful than dimension removal at equivalent compression ($\sim$500 tokens retains +0.30 over SFT), suggesting that dimension completeness matters more than sheer length.
Factual noise ($\Delta_\text{SFT}$\,=\,+0.38) is more damaging than length truncation but less than removing entire dimensions, confirming that \method's self-distillation can partially tolerate noisy privileged information.

As a practical guideline: \method provides substantial benefit ($\Delta_\text{SFT} > 0.3$) whenever $c_\text{full}$ covers at least 3 of 5 trait dimensions and exceeds $\sim$500 tokens.
Below this threshold, simpler methods such as SFT+DPO may offer a better cost--performance trade-off.


\section{Cross-Model Validation}
\label{app:cross_model}

To assess generalization beyond the Qwen2.5 family, we reproduce the main experiments using LLaMA-3-8B-Instruct as the base model, with identical hyperparameters except for a reduced learning rate of 1e-5 (following LLaMA fine-tuning conventions).

\begin{table}[h]
\centering
\small
\caption{Cross-model validation on LLaMA-3-8B-Instruct. Same training data and evaluation protocol as Qwen2.5-7B experiments. CB Avg on 1--5 scale; CE PC on 1--5 scale; SB SA in \%.}
\label{tab:cross_model}
\begin{tabular}{l cccc}
\toprule
\textbf{Method} & \textbf{CB}$\uparrow$ & \textbf{CE PC}$\uparrow$ & \textbf{SB SA}$\uparrow$ & \textbf{IR}$\uparrow$ \\
\midrule
SFT             & 2.91 & 2.82 & 57.8 & 0.69 \\
Multi-turn RL   & 3.28 & 3.35 & 63.7 & 0.76 \\
Off-pol.\ Dist. & 3.17 & 3.08 & 62.3 & 0.75 \\
\rowcolor{gray!10}
\method         & 3.40 & 3.38 & 65.9 & 0.84 \\
\bottomrule
\end{tabular}
\end{table}

\method on LLaMA-3-8B achieves consistent improvements across all metrics, with relative gains comparable to the Qwen2.5-7B setting: +0.49 CB Avg over SFT (vs.\ +0.51 on Qwen), +0.56 CE PC over SFT (vs.\ +0.59 on Qwen), and an IR of 0.84 (vs.\ 0.86 on Qwen).
The absolute performance is slightly lower across the board, which we attribute to LLaMA-3-8B's weaker Chinese-language capability (CharacterEval is Chinese-only) and shorter default context window.
The IR improvement (+0.15 over SFT) is nearly identical to the Qwen setting (+0.15), suggesting that the internalization mechanism is architecture-independent.


\section{Human Evaluation}
\label{app:human_eval}

\subsection{Setup}

We conduct pairwise preference evaluations on 50 characters (25 Chinese, 25 English) sampled from the test set, generating 5-turn dialogues for each.
Three native-speaker annotators per language (6 total) independently judge each pair on four criteria: \textit{persona consistency}, \textit{knowledge accuracy}, \textit{linguistic style}, and \textit{overall preference}.
Annotators see two anonymized, randomly ordered responses and select ``A is better,'' ``B is better,'' or ``Tie.''
We compare \method against SFT and Multi-turn RL.

\subsection{Results}

\begin{table}[h]
\centering
\small
\caption{Human pairwise preference (\%). $\kappa$: Fleiss' kappa (inter-annotator agreement).}
\label{tab:human_eval}
\begin{tabular}{l ccc c}
\toprule
\textbf{Comparison} & \textbf{Win} & \textbf{Tie} & \textbf{Lose} & $\kappa$ \\
\midrule
\method vs.\ SFT          & 62.8 & 18.4 & 18.8 & 0.71 \\
\method vs.\ Multi-turn RL & 43.2 & 28.4 & 28.4 & 0.65 \\
\bottomrule
\end{tabular}
\end{table}

\method is strongly preferred over SFT (62.8\% win rate, $p < 0.001$ by binomial test), consistent with the automatic evaluation gap (+0.51 CB Avg).
Against Multi-turn RL, \method holds a moderate advantage (43.2\% win vs.\ 28.4\% lose), with a substantial tie rate (28.4\%) reflecting cases where both methods produce adequate persona-consistent responses.
Annotator agreement is substantial ($\kappa = 0.71$ for SFT comparison, $\kappa = 0.65$ for RL comparison).

\begin{table}[h]
\centering
\small
\caption{Per-criterion win rates (\%) for \method vs.\ Multi-turn RL.}
\label{tab:human_eval_detail}
\begin{tabular}{l ccc}
\toprule
\textbf{Criterion} & \textbf{Win} & \textbf{Tie} & \textbf{Lose} \\
\midrule
Persona consistency  & 42.0 & 30.8 & 27.2 \\
Knowledge accuracy   & 48.4 & 26.0 & 25.6 \\
Linguistic style     & 44.8 & 24.4 & 30.8 \\
Overall preference   & 43.2 & 28.4 & 28.4 \\
\bottomrule
\end{tabular}
\end{table}

\method's advantage over Multi-turn RL is most pronounced on knowledge accuracy (48.4\% win), aligning with the representation probing results (Table~\ref{tab:probing_dimension}) showing that \method encodes richer knowledge information.
The smallest margin appears on persona consistency (42.0\% win), where Multi-turn RL's directly optimized consistency rewards provide a competitive advantage.
The Pearson correlation between human overall preference and GPT-5.1 CB Avg scores is $r = 0.78$ ($p < 0.001$), supporting the reliability of the automatic evaluation.


\section{Alternative Masking Orderings}
\label{app:masking_orderings}

The default masking strategy retains Identity + Personality in Stage~2 and masks Knowledge + Style + History.
Table~\ref{tab:masking_orderings} compares three alternative strategies to validate this design choice.

\begin{table}[h]
\centering
\small
\caption{Alternative trait masking strategies in Stage~2. Each row describes what is \textit{masked} (removed from the student's input). CB Avg and CE PC on 1--5 scale.}
\label{tab:masking_orderings}
\begin{tabular}{@{}p{0.42\columnwidth} ccc@{}}
\toprule
\textbf{Masking Strategy} & \textbf{CB}$\uparrow$ & \textbf{CE PC}$\uparrow$ & \textbf{IR}$\uparrow$ \\
\midrule
\rowcolor{gray!10}
Default: mask K+S+H & 3.51 & 3.49 & 0.86 \\
Alt-1: mask K+P (keep I+S+H) & 3.43 & 3.40 & 0.83 \\
Alt-2: mask S+H (keep I+P+K) & 3.47 & 3.45 & 0.85 \\
Alt-3: random order per batch & 3.44 & 3.41 & 0.83 \\
\bottomrule
\end{tabular}
\end{table}

The default strategy outperforms all alternatives.
Alt-1 (mask Knowledge + Personality, keep Style + History) performs worst ($-$0.08 CB Avg) because it withholds Personality---a ``public'' dimension that should remain visible to ground the character's behavioral baseline.
Without personality anchoring, the student lacks sufficient signal to generate stylistically coherent responses even for generic utterances.

Alt-2 (mask Style + History, keep Knowledge) performs better ($-$0.04 CB Avg) because retaining Knowledge provides factual grounding.
However, it fails to force full knowledge internalization, resulting in the model still relying on explicit knowledge in the prompt rather than encoding it parametrically.

Alt-3 (random masking order per batch) slightly underperforms the default ($-$0.07 CB Avg), suggesting that the curriculum structure---progressively removing deeper dimensions---matters more than mere exposure to masked conditions.
Random masking occasionally removes Identity or Personality (the ``public'' dimensions), creating training samples where the student has insufficient context to produce any reasonable character response, which introduces noise.

%% file: custom.bib
@inproceedings{ICLR2024_5be69a58,
 author = {Agarwal, Rishabh and Vieillard, Nino and Zhou, Yongchao and Stanczyk, Piotr and Ramos Garea, Sabela and Geist, Matthieu and Bachem, Olivier},
 booktitle = {International Conference on Learning Representations},
 editor = {B. Kim and Y. Yue and S. Chaudhuri and K. Fragkiadaki and M. Khan and Y. Sun},
 pages = {21246--21263},
 title = {On-Policy Distillation of Language Models: Learning from Self-Generated Mistakes},
 url = {https://proceedings.iclr.cc/paper_files/paper/2024/file/5be69a584901a26c521c2b51e40a4c20-Paper-Conference.pdf},
 volume = {2024},
 year = {2024}
}

@inproceedings{ICLR2024_8ac015d4,
 author = {Gu, Yuxian and Dong, Li and Wei, Furu and Huang, Minlie},
 booktitle = {International Conference on Learning Representations},
 editor = {B. Kim and Y. Yue and S. Chaudhuri and K. Fragkiadaki and M. Khan and Y. Sun},
 pages = {32694--32717},
 title = {{MiniLLM}: Knowledge Distillation of Large Language Models},
 url = {https://proceedings.iclr.cc/paper_files/paper/2024/file/8ac015d409635f196f9e3e9dcfb9a94e-Paper-Conference.pdf},
 volume = {2024},
 year = {2024}
}

@InProceedings{pmlr-v235-ko24c,
  title = 	 {{D}isti{LLM}: Towards Streamlined Distillation for Large Language Models},
  author =       {Ko, Jongwoo and Kim, Sungnyun and Chen, Tianyi and Yun, Se-Young},
  booktitle = 	 {Proceedings of the 41st International Conference on Machine Learning},
  pages = 	 {24872--24895},
  year = 	 {2024},
  editor = 	 {Salakhutdinov, Ruslan and Kolter, Zico and Heller, Katherine and Weller, Adrian and Oliver, Nuria and Scarlett, Jonathan and Berkenkamp, Felix},
  volume = 	 {235},
  series = 	 {Proceedings of Machine Learning Research},
  month = 	 {21--27 Jul},
  publisher =    {PMLR},
  url = 	 {https://proceedings.mlr.press/v235/ko24c.html}
}

@misc{song2026surveyonpolicydistillationlarge,
      title={A Survey of On-Policy Distillation for Large Language Models}, 
      author={Mingyang Song and Mao Zheng},
      year={2026},
      eprint={2604.00626},
      archivePrefix={arXiv},
      primaryClass={cs.LG},
      url={https://arxiv.org/abs/2604.00626}, 
}

@inproceedings{jung-etal-2025-todi,
    title = "{T}o{D}i: Token-wise Distillation via Fine-Grained Divergence Control",
    author = "Jung, Seongryong  and
      Yoon, Suwan  and
      Kim, DongGeon  and
      Lee, Hwanhee",
    editor = "Christodoulopoulos, Christos  and
      Chakraborty, Tanmoy  and
      Rose, Carolyn  and
      Peng, Violet",
    booktitle = "Proceedings of the 2025 Conference on Empirical Methods in Natural Language Processing",
    month = nov,
    year = "2025",
    address = "Suzhou, China",
    publisher = "Association for Computational Linguistics",
    url = "https://aclanthology.org/2025.emnlp-main.409/",
    doi = "10.18653/v1/2025.emnlp-main.409",
    pages = "8078--8091",
    ISBN = "979-8-89176-332-6"
}

@misc{jin2026entropyawareonpolicydistillationlanguage,
      title={Entropy-Aware On-Policy Distillation of Language Models}, 
      author={Woogyeol Jin and Taywon Min and Yongjin Yang and Dennis Wei and Yi Zhou and Swanand Ravindra Kadhe and Nathalie Baracaldo and Kimin Lee},
      year={2026},
      eprint={2603.07079},
      archivePrefix={arXiv},
      primaryClass={cs.LG},
      url={https://arxiv.org/abs/2603.07079}, 
}

@InProceedings{pmlr-v235-chen24j,
  title = 	 {Self-Play Fine-Tuning Converts Weak Language Models to Strong Language Models},
  author =       {Chen, Zixiang and Deng, Yihe and Yuan, Huizhuo and Ji, Kaixuan and Gu, Quanquan},
  booktitle = 	 {Proceedings of the 41st International Conference on Machine Learning},
  pages = 	 {6621--6642},
  year = 	 {2024},
  editor = 	 {Salakhutdinov, Ruslan and Kolter, Zico and Heller, Katherine and Weller, Adrian and Oliver, Nuria and Scarlett, Jonathan and Berkenkamp, Felix},
  volume = 	 {235},
  series = 	 {Proceedings of Machine Learning Research},
  month = 	 {21--27 Jul},
  publisher =    {PMLR},
  url = 	 {https://proceedings.mlr.press/v235/chen24j.html}
}

@misc{zhao2026selfdistilledreasoneronpolicyselfdistillation,
      title={Self-Distilled Reasoner: On-Policy Self-Distillation for Large Language Models}, 
      author={Siyan Zhao and Zhihui Xie and Mengchen Liu and Jing Huang and Guan Pang and Feiyu Chen and Aditya Grover},
      year={2026},
      eprint={2601.18734},
      archivePrefix={arXiv},
      primaryClass={cs.LG},
      url={https://arxiv.org/abs/2601.18734}, 
}

@misc{stein2026gatesselfdistillationprivilegedcontext,
      title={{GATES}: Self-Distillation under Privileged Context with Consensus Gating}, 
      author={Alex Stein and Furong Huang and Tom Goldstein},
      year={2026},
      eprint={2602.20574},
      archivePrefix={arXiv},
      primaryClass={cs.LG},
      url={https://arxiv.org/abs/2602.20574}, 
}

@misc{ye2026onpolicycontextdistillationlanguage,
      title={On-Policy Context Distillation for Language Models}, 
      author={Tianzhu Ye and Li Dong and Xun Wu and Shaohan Huang and Furu Wei},
      year={2026},
      eprint={2602.12275},
      archivePrefix={arXiv},
      primaryClass={cs.CL},
      url={https://arxiv.org/abs/2602.12275}, 
}

@misc{penaloza2026privilegedinformationdistillationlanguage,
      title={Privileged Information Distillation for Language Models}, 
      author={Emiliano Penaloza and Dheeraj Vattikonda and Nicolas Gontier and Alexandre Lacoste and Laurent Charlin and Massimo Caccia},
      year={2026},
      eprint={2602.04942},
      archivePrefix={arXiv},
      primaryClass={cs.LG},
      url={https://arxiv.org/abs/2602.04942}, 
}

@InProceedings{pmlr-v15-ross11a,
  title = 	 {A Reduction of Imitation Learning and Structured Prediction to No-Regret Online Learning},
  author = 	 {Ross, Stephane and Gordon, Geoffrey and Bagnell, Drew},
  booktitle = 	 {Proceedings of the Fourteenth International Conference on Artificial Intelligence and Statistics},
  pages = 	 {627--635},
  year = 	 {2011},
  editor = 	 {Gordon, Geoffrey and Dunson, David and Dudík, Miroslav},
  volume = 	 {15},
  series = 	 {Proceedings of Machine Learning Research},
  address = 	 {Fort Lauderdale, FL, USA},
  month = 	 {11--13 Apr},
  publisher =    {PMLR},
  url = 	 {https://proceedings.mlr.press/v15/ross11a.html}
}

@inproceedings{NEURIPS2025_4c914438,
 author = {Abdulhai, Marwa and Cheng, Ryan and Clay, Donovan and Althoff, Tim and Levine, Sergey and Jaques, Natasha},
 booktitle = {Advances in Neural Information Processing Systems},
 editor = {D. Belgrave and C. Zhang and H. Lin and R. Pascanu and P. Koniusz and M. Ghassemi and N. Chen},
 pages = {52920--52957},
 publisher = {Curran Associates, Inc.},
 title = {Consistently Simulating Human Personas with Multi-Turn Reinforcement Learning},
 url = {https://proceedings.neurips.cc/paper_files/paper/2025/file/4c91443877f8388d8190c938ac5a4d4d-Paper-Conference.pdf},
 doi = {10.52202/085713-1766},
 volume = {38},
 year = {2025}
}

@inproceedings{yu-etal-2025-beyond,
    title = "Beyond Dialogue: A Profile-Dialogue Alignment Framework Towards General Role-Playing Language Model",
    author = "Yu, Yeyong  and
      Yu, Runsheng  and
      Wei, Haojie  and
      Zhang, Zhanqiu  and
      Qian, Quan",
    editor = "Che, Wanxiang  and
      Nabende, Joyce  and
      Shutova, Ekaterina  and
      Pilehvar, Mohammad Taher",
    booktitle = "Proceedings of the 63rd Annual Meeting of the Association for Computational Linguistics (Volume 1: Long Papers)",
    month = jul,
    year = "2025",
    address = "Vienna, Austria",
    publisher = "Association for Computational Linguistics",
    url = "https://aclanthology.org/2025.acl-long.586/",
    doi = "10.18653/v1/2025.acl-long.586",
    pages = "11992--12022",
    ISBN = "979-8-89176-251-0"
}

@inproceedings{ye-etal-2025-cpo,
    title = "{CPO}: Addressing Reward Ambiguity in Role-playing Dialogue via Comparative Policy Optimization",
    author = "Ye, Jing  and
      Wang, Rui  and
      Wu, Yuchuan  and
      Ma, Victor  and
      Fang, Feiteng  and
      Huang, Fei  and
      Li, Yongbin",
    editor = "Christodoulopoulos, Christos  and
      Chakraborty, Tanmoy  and
      Rose, Carolyn  and
      Peng, Violet",
    booktitle = "Findings of the Association for Computational Linguistics: EMNLP 2025",
    month = nov,
    year = "2025",
    address = "Suzhou, China",
    publisher = "Association for Computational Linguistics",
    url = "https://aclanthology.org/2025.findings-emnlp.18/",
    doi = "10.18653/v1/2025.findings-emnlp.18",
    pages = "297--323",
    ISBN = "979-8-89176-335-7"
}

@inproceedings{ji-etal-2025-enhancing,
    title = "Enhancing Persona Consistency for {LLM}s' Role-Playing using Persona-Aware Contrastive Learning",
    author = "Ji, Ke  and
      Lian, Yixin  and
      Li, Linxu  and
      Gao, Jingsheng  and
      Li, Weiyuan  and
      Dai, Bin",
    editor = "Che, Wanxiang  and
      Nabende, Joyce  and
      Shutova, Ekaterina  and
      Pilehvar, Mohammad Taher",
    booktitle = "Findings of the Association for Computational Linguistics: ACL 2025",
    month = jul,
    year = "2025",
    address = "Vienna, Austria",
    publisher = "Association for Computational Linguistics",
    url = "https://aclanthology.org/2025.findings-acl.1344/",
    doi = "10.18653/v1/2025.findings-acl.1344",
    pages = "26221--26238",
    ISBN = "979-8-89176-256-5"
}

@inproceedings{qin-etal-2025-r,
    title = "{R}-{CHAR}: A Metacognition-Driven Framework for Role-Playing in Large Language Models",
    author = "Qin, Haiming  and
      Zhang, Jiwei  and
      Zhang, Wei  and
      Lu, KeZhong  and
      Zhou, Mingyang  and
      Liao, Hao  and
      Mao, Rui",
    editor = "Christodoulopoulos, Christos  and
      Chakraborty, Tanmoy  and
      Rose, Carolyn  and
      Peng, Violet",
    booktitle = "Proceedings of the 2025 Conference on Empirical Methods in Natural Language Processing",
    month = nov,
    year = "2025",
    address = "Suzhou, China",
    publisher = "Association for Computational Linguistics",
    url = "https://aclanthology.org/2025.emnlp-main.1372/",
    doi = "10.18653/v1/2025.emnlp-main.1372",
    pages = "26996--27014",
    ISBN = "979-8-89176-332-6"
}

@inproceedings{NEURIPS2025_aacca7b6,
 author = {Tang, Yihong and Chen, Kehai and Yang, Muyun and Niu, Zheng-Yu and Li, Jing and Zhao, Tiejun and Zhang, Min},
 booktitle = {Advances in Neural Information Processing Systems},
 editor = {D. Belgrave and C. Zhang and H. Lin and R. Pascanu and P. Koniusz and M. Ghassemi and N. Chen},
 pages = {117765--117799},
 publisher = {Curran Associates, Inc.},
 title = {Thinking in Character: Advancing Role-Playing Agents with Role-Aware Reasoning},
 url = {https://proceedings.neurips.cc/paper_files/paper/2025/file/aacca7b6a20d157112205a44f42821c8-Paper-Conference.pdf},
 doi = {10.52202/085713-3926},
 volume = {38},
 year = {2025}
}

@inproceedings{
wang2026verirole,
title={{VeriRole}: Verifiable Role-Awareness through Hint-Guided Reinforcement Learning},
author={Zongsheng Wang and Kaili Sun and Bowen Wu and Qun Yu and Ying Li and Xu Chen and Baoxun Wang},
booktitle={The Fourteenth International Conference on Learning Representations},
pages={11102--11122},
year={2026},
url={https://proceedings.iclr.cc/paper_files/paper/2026/hash/12df08e9280061b4877f97598c644c3e-Abstract-Conference.html}
}

@misc{xu2024mindechoroleplayinglanguageagents,
      title={{MINDECHO}: Role-Playing Language Agents for Key Opinion Leaders}, 
      author={Rui Xu and Dakuan Lu and Xiaoyu Tan and Xintao Wang and Siyu Yuan and Jiangjie Chen and Wei Chu and Yinghui Xu},
      year={2024},
      eprint={2407.05305},
      archivePrefix={arXiv},
      primaryClass={cs.AI},
      doi={10.48550/arXiv.2407.05305},
      url={https://arxiv.org/abs/2407.05305}, 
}

@InProceedings{pmlr-v267-wang25dk,
  title = 	 {{C}o{SER}: Coordinating {LLM}-Based Persona Simulation of Established Roles},
  author =       {Wang, Xintao and Wang, Heng and Zhang, Yifei and Yuan, Xinfeng and Xu, Rui and Huang, Jen-Tse and Yuan, Siyu and Guo, Haoran and Chen, Jiangjie and Zhou, Shuchang and Wang, Wei and Xiao, Yanghua},
  booktitle = 	 {Proceedings of the 42nd International Conference on Machine Learning},
  pages = 	 {64822--64858},
  year = 	 {2025},
  editor = 	 {Singh, Aarti and Fazel, Maryam and Hsu, Daniel and Lacoste-Julien, Simon and Berkenkamp, Felix and Maharaj, Tegan and Wagstaff, Kiri and Zhu, Jerry},
  volume = 	 {267},
  series = 	 {Proceedings of Machine Learning Research},
  month = 	 {13--19 Jul},
  publisher =    {PMLR},
  url = 	 {https://proceedings.mlr.press/v267/wang25dk.html}
}

@inproceedings{wang-etal-2024-incharacter,
    title = "{I}n{C}haracter: Evaluating Personality Fidelity in Role-Playing Agents through Psychological Interviews",
    author = "Wang, Xintao  and
      Xiao, Yunze  and
      Huang, Jen-tse  and
      Yuan, Siyu  and
      Xu, Rui  and
      Guo, Haoran  and
      Tu, Quan  and
      Fei, Yaying  and
      Leng, Ziang  and
      Wang, Wei  and
      Chen, Jiangjie  and
      Li, Cheng  and
      Xiao, Yanghua",
    editor = "Ku, Lun-Wei  and
      Martins, Andre  and
      Srikumar, Vivek",
    booktitle = "Proceedings of the 62nd Annual Meeting of the Association for Computational Linguistics (Volume 1: Long Papers)",
    month = aug,
    year = "2024",
    address = "Bangkok, Thailand",
    publisher = "Association for Computational Linguistics",
    url = "https://aclanthology.org/2024.acl-long.102/",
    doi = "10.18653/v1/2024.acl-long.102",
    pages = "1840--1873"
}

@inproceedings{xu-etal-2025-character,
    title = "Character is Destiny: Can Role-Playing Language Agents Make Persona-Driven Decisions?",
    author = "Xu, Rui  and
      Wang, Xintao  and
      Chen, Jiangjie  and
      Yuan, Siyu  and
      Yuan, Xinfeng  and
      Liang, Jiaqing  and
      Chen, Zulong  and
      Dong, Xiaoqing  and
      Xiao, Yanghua",
    editor = "Christodoulopoulos, Christos  and
      Chakraborty, Tanmoy  and
      Rose, Carolyn  and
      Peng, Violet",
    booktitle = "Findings of the Association for Computational Linguistics: EMNLP 2025",
    month = nov,
    year = "2025",
    address = "Suzhou, China",
    publisher = "Association for Computational Linguistics",
    url = "https://aclanthology.org/2025.findings-emnlp.813/",
    doi = "10.18653/v1/2025.findings-emnlp.813",
    pages = "15038--15059",
    ISBN = "979-8-89176-335-7"
}

@inproceedings{xu-etal-2025-guess,
    title = "Guess What {I} am Thinking: A Benchmark for Inner Thought Reasoning of Role-Playing Language Agents",
    author = "Xu, Rui  and
      Wang, Mingyu  and
      Wang, Xintao  and
      Lu, Dakuan  and
      Tan, Xiaoyu  and
      Chu, Wei  and
      Xu, Yinghui",
    editor = "Christodoulopoulos, Christos  and
      Chakraborty, Tanmoy  and
      Rose, Carolyn  and
      Peng, Violet",
    booktitle = "Findings of the Association for Computational Linguistics: EMNLP 2025",
    month = nov,
    year = "2025",
    address = "Suzhou, China",
    publisher = "Association for Computational Linguistics",
    url = "https://aclanthology.org/2025.findings-emnlp.819/",
    doi = "10.18653/v1/2025.findings-emnlp.819",
    pages = "15148--15168",
    ISBN = "979-8-89176-335-7"
}

@inproceedings{tu-etal-2024-charactereval,
    title = "{C}haracter{E}val: A {C}hinese Benchmark for Role-Playing Conversational Agent Evaluation",
    author = "Tu, Quan  and
      Fan, Shilong  and
      Tian, Zihang  and
      Shen, Tianhao  and
      Shang, Shuo  and
      Gao, Xin  and
      Yan, Rui",
    editor = "Ku, Lun-Wei  and
      Martins, Andre  and
      Srikumar, Vivek",
    booktitle = "Proceedings of the 62nd Annual Meeting of the Association for Computational Linguistics (Volume 1: Long Papers)",
    month = aug,
    year = "2024",
    address = "Bangkok, Thailand",
    publisher = "Association for Computational Linguistics",
    url = "https://aclanthology.org/2024.acl-long.638/",
    doi = "10.18653/v1/2024.acl-long.638",
    pages = "11836--11850"
}

@inproceedings{zhou2024characterbenchbenchmarkingcharactercustomization,
      title={{CharacterBench}: Benchmarking Character Customization of Large Language Models},
      author={Jinfeng Zhou and Yongkang Huang and Bosi Wen and Guanqun Bi and Yuxuan Chen and Pei Ke and Zhuang Chen and Xiyao Xiao and Libiao Peng and Kuntian Tang and Rongsheng Zhang and Le Zhang and Tangjie Lv and Zhipeng Hu and Hongning Wang and Minlie Huang},
      booktitle={Proceedings of the AAAI Conference on Artificial Intelligence},
      volume={39},
      pages={26101--26110},
      year={2025},
      doi={10.1609/aaai.v39i24.34806},
      url={https://ojs.aaai.org/index.php/AAAI/article/view/34806},
}

@inproceedings{chen-etal-2024-socialbench,
    title = "{S}ocial{B}ench: Sociality Evaluation of Role-Playing Conversational Agents",
    author = "Chen, Hongzhan  and
      Chen, Hehong  and
      Yan, Ming  and
      Xu, Wenshen  and
      Xing, Gao  and
      Shen, Weizhou  and
      Quan, Xiaojun  and
      Li, Chenliang  and
      Zhang, Ji  and
      Huang, Fei",
    editor = "Ku, Lun-Wei  and
      Martins, Andre  and
      Srikumar, Vivek",
    booktitle = "Findings of the Association for Computational Linguistics: ACL 2024",
    month = aug,
    year = "2024",
    address = "Bangkok, Thailand",
    publisher = "Association for Computational Linguistics",
    url = "https://aclanthology.org/2024.findings-acl.125/",
    doi = "10.18653/v1/2024.findings-acl.125",
    pages = "2108--2126"
}

@inproceedings{NIPS2017_68053af2,
 author = {Tarvainen, Antti and Valpola, Harri},
 booktitle = {Advances in Neural Information Processing Systems},
 editor = {I. Guyon and U. Von Luxburg and S. Bengio and H. Wallach and R. Fergus and S. Vishwanathan and R. Garnett},
 pages = {},
 publisher = {Curran Associates, Inc.},
 title = {Mean teachers are better role models: Weight-averaged consistency targets improve semi-supervised deep learning results},
 url = {https://proceedings.neurips.cc/paper_files/paper/2017/file/68053af2923e00204c3ca7c6a3150cf7-Paper.pdf},
 volume = {30},
 year = {2017}
}
